\documentclass{article}

\usepackage{microtype}
\usepackage{graphicx}
\usepackage{subcaption}
\usepackage{booktabs} 
\usepackage{multirow}

\usepackage{hyperref}

\usepackage[preprint]{icml2026}

\usepackage{amsmath}
\usepackage{amssymb}
\usepackage{mathtools}
\usepackage{amsthm}

\usepackage[capitalize,noabbrev]{cleveref}

\theoremstyle{plain}

\theoremstyle{definition}

\theoremstyle{remark}

\newcommand{\bsdisable}{}

\usepackage[textsize=tiny]{todonotes}

\icmltitlerunning{FlowBypass: Rectified Flow Trajectory Bypass for Training-Free Image Editing}

\begin{document}

\twocolumn[
  \icmltitle{FlowBypass: Rectified Flow Trajectory Bypass for Training-Free Image Editing}



  \icmlsetsymbol{equal}{*}

  \begin{icmlauthorlist}
    \icmlauthor{Menglin Han}{tj}
    \icmlauthor{Zhangkai Ni}{tj}
  \end{icmlauthorlist}

  \icmlaffiliation{tj}{School of Computer Science and Technology, Tongji University, Shanghai, China}
\icmlcorrespondingauthor{Zhangkai Ni}{zkni@tongji.edu.cn}

  \icmlkeywords{Image Editing, Rectified Flow, Artificial Intelligence Generated Content}

  \vskip 0.3in
]



\printAffiliationsAndNotice{}  

\begin{abstract}
Training-free image editing has attracted increasing attention for its efficiency and independence from training data. 
However, existing approaches predominantly rely on inversion–reconstruction trajectories, which impose an inherent trade-off: longer trajectories accumulate errors and compromise fidelity, while shorter ones fail to ensure sufficient alignment with the edit prompt. 
Previous attempts to address this issue typically employ backbone-specific feature manipulations, limiting general applicability. 
To address these challenges, we propose FlowBypass, a novel and analytical framework grounded in Rectified Flow that constructs a bypass directly connecting inversion and reconstruction trajectories, thereby mitigating error accumulation without relying on feature manipulations. 
We provide a formal derivation of two trajectories, from which we obtain an approximate bypass formulation and its numerical solution, enabling seamless trajectory transitions. 
Extensive experiments demonstrate that FlowBypass consistently outperforms state-of-the-art image editing methods, achieving stronger prompt alignment while preserving high-fidelity details in irrelevant regions.
\end{abstract}
\section{Introduction}
Image editing has become a powerful paradigm for controlling and manipulating visual content through intuitive instructions and prompts. 
Among various approaches, \textit{training-free image editing} has gained increasing attention because it avoids the need for large-scale data or costly fine-tuning while still enabling flexible and effective manipulations. 
This property makes training-free methods especially appealing for practical deployment across diverse real-world applications.

However, a persistent challenge limits the effectiveness of existing training-free methods. 
Most existing approaches are built upon inversion–reconstruction trajectories, where an image is first inverted into noise and then reconstructed under the guidance of an edit prompt. 
This design inherently creates a trade-off: long trajectories accumulate estimation errors that erode \textbf{fidelity} in regions that should remain untouched, while short trajectories weaken \textbf{alignment} with the prompt, resulting in incomplete edits. 
Prior efforts to alleviate this trade-off include trajectory adjustments~\citep{brack2024ledits, rout2024semantic}, prompt refinements~\citep{mokady2023null, miyake2025negative}, and feature manipulations~\citep{Hertz2023PrompttoPromptIE,simsar2025lime}. 
However, most of these methods traverse full trajectories, which exacerbates error accumulation near the inverted noise, and some rely on backbone-specific feature interventions, which hinder generality across generative models.

To overcome these limitations, we introduce \textbf{FlowBypass}, a general and analytical training-free image editing framework grounded in Rectified Flow (RF)~\citep{liu2022flow}. 
The central idea is a \textbf{trajectory bypass} that directly connects the inversion and reconstruction trajectories at intermediate states, thereby reducing accumulated estimation errors without relying on feature manipulations. 
This design not only addresses fidelity–alignment trade-offs but also ensures strong generalizability across diverse backbones. 
We develop FlowBypass starting from theoretical insights: we first analyze the inversion and reconstruction trajectories, then derive an approximate formulation of the bypass, and finally design a numerical solution via Euler discretization.
This theoretical-to-practical pipeline enables efficient trajectory transitions while providing a unified solution applicable to a wide class of Rectified Flow models.
Extensive experiments across diverse editing tasks demonstrate that FlowBypass consistently outperforms state-of-the-art training-free methods, achieving stronger alignment with edit prompts while preserving high-fidelity details in irrelevant regions.
Our main contributions are summarized as follows:

$\bullet$ \textbf{Theoretical Foundation of FlowBypass}: We provide a rigorous mathematical formulation of inversion and reconstruction trajectories, and derive an approximate bypass solution with a tractable analytical form, laying the theoretical groundwork for FlowBypass.

$\bullet$ \textbf{Unified and Practical Realization}: We transform the analytical solution into an efficient discretized form, yielding a unified training-free image editing framework based on Rectified Flow. It bridges theory and practice by bypassing inversion to directly guide reconstruction, removing the need for backbone-specific feature manipulations and improving both fidelity and alignment.

$\bullet$ \textbf{State-of-the-Art Performance}: Extensive experiments demonstrate that FlowBypass consistently achieves state-of-the-art results across challenging editing scenarios, delivering superior fidelity–alignment trade-offs and robust generalization compared to existing training-free methods.

\section{Related Works}
Training-free image editing aims to manipulate images without additional training or fine-tuning. Rather than updating model parameters~\citep{brooks2023instructpix2pix,yu2025anyedit,chen2025unireal,xiao2025omnigen}, these methods directly exploit pre-trained generative models. 
Existing approaches can be broadly grouped into {four} categories according to how they modify the sampling process.  

The most common paradigm is noise-inversion methods, the dominant paradigm where an original image is mapped into the noise space and then reconstructed into the edited result. 
Most build upon DDIM-inversion~\citep{Song2021DenoisingDI}, with modifications to inversion and reconstruction trajectories~\citep{HubermanSpiegelglas2023AnEF,mokady2023null,brack2024ledits,miyake2025negative}, but they remain fundamentally constrained by accumulated errors, especially near the inverted noise.  
Recently, training-free editing has also been extended to Rectified Flow (RF) models~\citep{wang2024taming,rout2024semantic}, but most remain rooted in DDIM-based principles, which continue to suffer from the accumulation of trajectory errors.

The second paradigm is prompt- or condition-refinement, which enhances the conditioning signal by incorporating information from the source image, thereby enabling edits within the reconstruction trajectory~\citep{Ravi2023PRedItORTG,Wang2023InstructEditIA}. 
The third paradigm is feature manipulation, where intermediate representations are modified either by injecting features from the inversion trajectory to improve fidelity or by amplifying prompt-related activations to strengthen alignment~\citep{Hertz2023PrompttoPromptIE,tumanyan2023plug,Cao2023MasaCtrlTM,feng2025dit4edit}. 

The fourth paradigm skips explicit inversion, using a coarse preliminary interception for pseudo-inversion and then operating directly in the image distribution, which is far more complex than the noise-blended domain~\citep{xu2024inversion,kulikov2025flowedit}. Consequently, inversion-free methods may introduce artifacts during editing, often resulting in degraded visual quality.

Despite their methodological diversity, most of these methods rely on full trajectory traversal, which inherently amplifies discretization errors and weak conditional guidance. 
As a result, training-free editing often introduces unintended modifications in irrelevant regions or even fails to realize the intended edit. 
This persistent bottleneck motivates the development of \textbf{FlowBypass}, a principled framework designed to mitigate accumulated errors while preserving both fidelity and alignment.

\section{Method}

\begin{figure*}[t]
    \centering
    \includegraphics[width=0.92\linewidth]{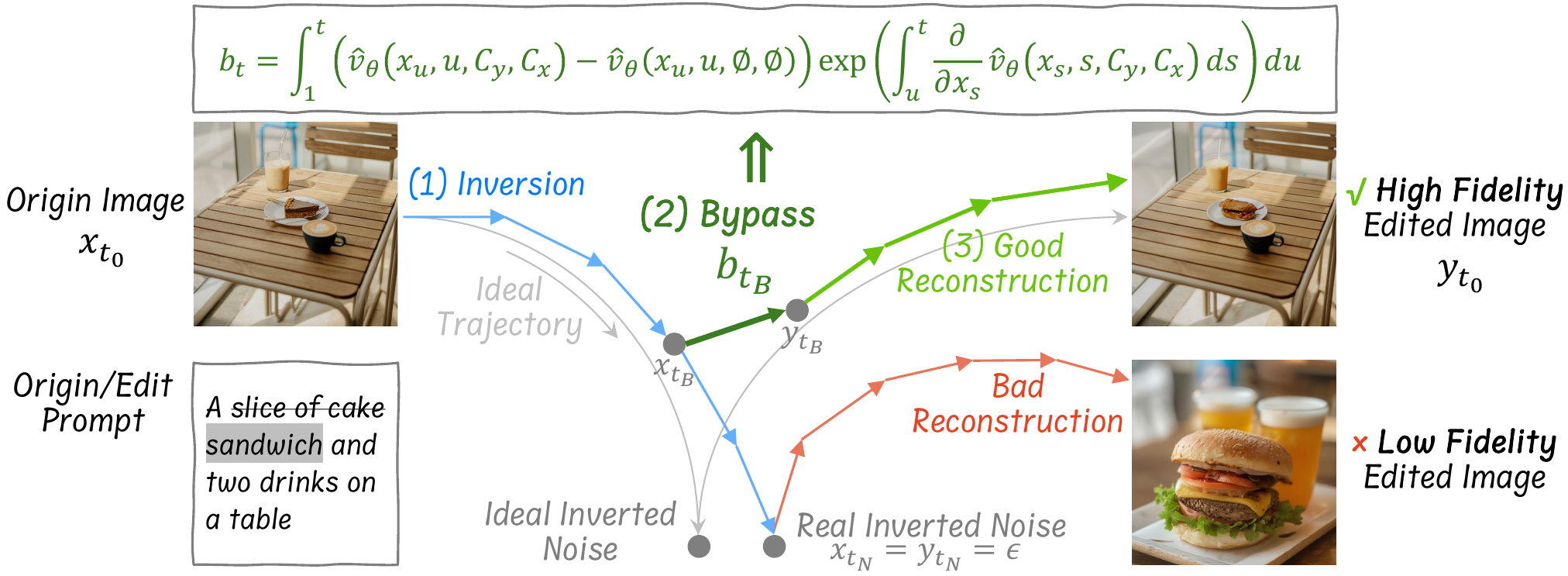}
    \caption{
    \textbf{Framework of FlowBypass.} FlowBypass consists of three steps: \textbf{(1)} Inverse the input image ${x}_{t_0}$ with Euler discretization to obtain the inversion trajectory; \textbf{(2)} Calculate the bypass $\bsdisable{b}_{t_B}$ using inversion trajectory according to Equ.~\ref{equ:discret_form}; \textbf{(3)} Reconstruct from intermediate state $\bsdisable{y}_{t_B} = \bsdisable{x}_{t_B} + \bsdisable{b}_{t_B}$ to obtain edited image $\bsdisable{y}_{t_0}$.
    We show the differences between the origin prompt and the edit prompt by marking removed words with strikethrough and added words with a gray background.}
    \label{fig:framework}
\end{figure*}

\subsection{Preliminaries}
\textbf{Rectified Flow (RF).}
RF formulates image generation as a continuous transformation of a Gaussian noise distribution $\mathcal{N}(0,\boldsymbol{I})$ into the target data distribution $p_{\text{data}}$ under conditioning signals, governed by a velocity field expressed as an ordinary differential equation (ODE):
\begin{equation}
    \frac{d}{dt}\bsdisable{z}_t = \bar{v}(\bsdisable{z}_t, t, C),
    \label{equ:d_ode}
\end{equation}
where $t \in [0, 1]$ is the continuous timestep, $\bsdisable{z}_0 \sim p_{\text{data}}$ is a real image, $\bsdisable{z}_1 = \bsdisable{\epsilon} \sim \mathcal{N}(0,\boldsymbol{I})$ is a Gaussian noise sample, $\bar{v}$ is the marginal velocity field~\citep{lipman2023flow}, and $C$ is the provided condition (\textit{e.g.}, text prompts).
The RF network $v_{\theta}$ is trained to approximate the marginal velocity field $\bar{v}$ by minimizing $\mathbb{E}_{t,\bsdisable{z}_0,\bsdisable{\epsilon}}\Vert v_{\theta}(\bsdisable{z}_t, t, C) - (\bsdisable{\epsilon} - \bsdisable{z}_0) \Vert^2$, where ${ \bsdisable{z}_t = t \cdot \epsilon + (1-t) \cdot z_0 }$ is the linear interpolation of $\bsdisable{z}_0$ and $\bsdisable{\epsilon}$.
Sampling seeks to recover a clean image $\bsdisable{z}_0$ from noise $\bsdisable{\epsilon}$ using Euler discretization of Equ.~\ref{equ:d_ode}, once the velocity field is well estimated by $v_\theta$.

\textbf{Classifier-Free Guidance (CFG).} 
CFG is widely used to balance semantic alignment and visual fidelity during sampling. 
Given velocity fields predicted under a positive prompt $C_p$ and a negative prompt $C_n$, the guided velocity field can be formulated as:
\begin{equation}
  \hat{v}_{\theta}(\bsdisable{z}_t, t, C_p, C_n) = v_{\theta}(\bsdisable{z}_t, t, C_n) + \omega \cdot \Big(v_{\theta}(\bsdisable{z}_t, t, C_p) - v_{\theta}(\bsdisable{z}_t, t, C_n) \Big),
\label{equ:cfg_define}
\end{equation}
where $\omega$ is the guidance scale.
The larger values of $\omega$ amplify the contribution of positive prompt, typically leading to improved semantic alignment with provided condition but at the risk of sacrificing diversity and visual fidelity.
Substituting $\hat{v}_{\theta}$ into Equ.~\ref{equ:d_ode} yields the CFG-guided trajectory:
\begin{equation}
  \bsdisable{z}_t = \bsdisable{z}_r + \int_r^t \hat{v}_{\theta}(\bsdisable{z}_{\tau}, \tau, C_p, C_n) d\tau.
  \label{equ:int_transfer}
\end{equation}

\textbf{Numerical Integration.} 
The above ODEs can be solved numerically through discretization. 
Starting from $\bsdisable{z}_{t_N}=\bsdisable{\epsilon}\!\sim\!\mathcal{N}(0,\boldsymbol{I})$ at $t_N=1$ and using a monotonic sequence of timesteps $\{t_0=0, t_1, \dots, t_N=1\}$, Euler discretization updates the state as:
\begin{equation}
    \bsdisable{z}_{t_{i-1}} = \bsdisable{z}_{t_i} + (t_{i-1} - t_i) \cdot \hat{v}_{\theta}(\bsdisable{z}_{t_i}, t_i, C_p, C_n).
    \label{equ:dis_ode}
\end{equation}

Iteratively applying Equ.~\ref{equ:dis_ode}, the initial noise gradually moves along the approximated trajectory until it reaches the final clean image $\bsdisable{z}_{t_0}$.

\textbf{Inversion-based Image Editing.}
Given a pre-trained generative model $\mathcal{M}$, an input image $\bsdisable{x}_{t_0}$, and an origin condition $C_x$, the objective is to synthesize an edited image $\bsdisable{y}_{t_0}$ that satisfies a user-specified edit condition $C_y$, while preserving the original content not related to the edit condition. 
The dominant pipeline consists of two stages: (i) \emph{inversion}, mapping $\bsdisable{x}_{t_0}$ into a corresponding Gaussian noise $\bsdisable{x}_{t_N}\equiv\bsdisable{\epsilon}\equiv\bsdisable{y}_{t_N}$ under the guidance of inversion condition $C_{inv}$; and (ii) \emph{reconstruction}, resampling from the noise $\bsdisable{y}_{t_N}$ to reconstruct the edited image $\bsdisable{y}_{t_0}$ under the guidance of reconstruction condition $C_{rec}$.
{The inversion and reconstruction trajectories share the same noise point $\bsdisable{x}_{t_N}\equiv\bsdisable{\epsilon}\equiv\bsdisable{y}_{t_N}$, while differing in their image points $x_0$ (origin image) and $y_0$ (edited image).}
While the design of $C_{inv}$ and $C_{rec}$ varies across methods, all such approaches depend on traversing the full inversion–reconstruction trajectory. 
This reliance makes them inherently vulnerable to numerical error accumulation and attenuated conditional guidance, which motivates the bypass strategy in our framework.

\subsection{Motivation}
A fundamental challenge in inversion–based training-free image editing is the progressive accumulation of errors during inversion and reconstruction. 
These errors in inversion primarily stem from two sources: (i) numerical discretization when solving the ODE in Equ.~\ref{equ:dis_ode}, and (ii) intrinsic mismatch between inversion and forward sampling. 
Concretely, the inversion equation $\bsdisable{x}_{t_i} = \bsdisable{x}_{t_{i-1}} + (t_i - t_{i-1})\cdot\hat{v}_{\theta}(\bsdisable{x}_{t_i}, t_i, C_p, C_n)$ reformulated from Equ.~\ref{equ:dis_ode} contains $\bsdisable{x}_{t_i}$ on both sides, thereby necessitating an approximation of $\hat{v}_{\theta}(\bsdisable{x}_{t_i}, t_i, C_p, C_n)$ with either $\hat{v}_{\theta}(\bsdisable{x}_{t_{i-1}}, t_i, C_p, C_n)$ or $\hat{v}_{\theta}(\bsdisable{x}_{t_{i-1}}, t_{i-1}, C_p, C_n)$. 
Such approximation introduces errors that propagate along the trajectory, ultimately yielding a degraded terminal state $\bsdisable{x}_{t_N}$. 
While the reconstruction trajectory is unaffected by intrinsic mismatch, it is still subject to errors caused by numerical discretization.

Our key insight is that fidelity loss in training-free editing is largely attributable to reconstructing from this corrupted terminal noise $\bsdisable{x}_{t_N}$. 
This observation motivates a departure from conventional designs: rather than relying on the error-prone endpoint, we propose to bypass it by selecting an intermediate state $\bsdisable{x}_{t_B}$, which preserves higher fidelity due to reduced error accumulation. 
We then construct the edited counterpart $\bsdisable{y}_{t_B}$ by introducing a bypass term $\bsdisable{b}_{t_B}$, as illustrated in Fig.~\ref{fig:framework}. 
This bypass enables a direct transition from inversion to reconstruction at $t_B$, effectively mitigating accumulated errors while ensuring semantic alignment with the edit prompt through $\bsdisable{b}_{t_B}$.

\subsection{Formulation of FlowBypass}
Given a pre-trained generative RF network $v_{\theta}$, an origin image $\bsdisable{x}_0$ with its associated origin prompt $C_x$, and a target edit prompt $C_y$ specifying the desired output $\bsdisable{y}_0$, two ODEs can be established as: 
\begin{equation}
    \begin{split}
        \bsdisable{x}_t &= \bsdisable{x}_1 + \int_1^t \hat{v}_{\theta}(\bsdisable{x}_{\tau}, \tau, C_{inv}^p, C_{inv}^n) d\tau, \\
        \bsdisable{y}_t &= \bsdisable{y}_1 + \int_1^t \hat{v}_{\theta}(\bsdisable{y}_{\tau}, \tau, C_{rec}^p, C_{rec}^n) d\tau, \\
    \end{split}
\end{equation}
where $C_{inv}^p$ and $C_{inv}^n$ are the positive and negative prompts that control the inversion trajectory, and $C_{rec}^p$ and $C_{rec}^n$ control the reconstruction trajectory.
The specific choice of these prompts is detailed in Sec.~\ref{sec:flowbypass_implement}.
We define the bypass $\bsdisable{b}_t$ as the offset between the two trajectories:
\begin{equation}
    \begin{split}
        \bsdisable{b}_t &= \bsdisable{y}_t - \bsdisable{x}_t \\
        &= \int_1^t \Big(\hat{v}_{\theta}(\bsdisable{y}_{\tau}, \tau, C_{rec}^p, C_{rec}^n) - \hat{v}_{\theta}(\bsdisable{x}_{\tau}, \tau, C_{inv}^p, C_{inv}^n) \Big) d\tau.
    \end{split}
\end{equation}
Differentiating both sides with respect to $t$, which leads to:
\begin{equation}
    \begin{split}
        \frac{d}{dt}\bsdisable{b}_t &= \hat{v}_{\theta}(\bsdisable{y}_{t}, t, C_{rec}^p, C_{rec}^n) - \hat{v}_{\theta}(\bsdisable{x}_{t}, t, C_{inv}^p, C_{inv}^n) \\
        &= \hat{v}_{\theta}(\bsdisable{x}_{t} + \bsdisable{b}_{t}, t, C_{rec}^p, C_{rec}^n) - \hat{v}_{\theta}(\bsdisable{x}_{t}, t, C_{inv}^p, C_{inv}^n). \\
    \end{split}
    \label{equ:ode_b_t}
\end{equation}

Since $\bsdisable{b}_{t}$ can be regarded as a small offset, so that the first term $\hat{v}_{\theta}(\bsdisable{x}_{t} + \bsdisable{b}_{t}, t, C_{rec}^p, C_{rec}^n)$ can be approximated with First-Order Taylor Expansion Formula:
\begin{equation}
    \begin{split}
        \hat{v}_{\theta}(\bsdisable{x}_{t} + \bsdisable{b}_{t}, t, C_{rec}^p, C_{rec}^n) &\approx \hat{v}_{\theta}(\bsdisable{x}_{t}, t, C_{rec}^p, C_{rec}^n) + P_t \cdot \bsdisable{b}_{t}, \\
        P_t &= \frac{\partial}{\partial \bsdisable{x}_{t}}\hat{v}_{\theta}(\bsdisable{x}_{t}, t, C_{rec}^p, C_{rec}^n) . \\
    \end{split}
    \label{equ:v_theta_approx}
\end{equation}

Substituting Equ.~\ref{equ:v_theta_approx} into Equ.~\ref{equ:ode_b_t} yields an approximate linear ODE for $\bsdisable{b}_t$:
\begin{equation}
  \begin{split}
    \bsdisable{b}_t &\approx \bsdisable{b}^*_t,\bsdisable{b}^*_1 = \bsdisable{0}, \\
    \frac{d}{dt}\bsdisable{b}^*_t &= Q_t + P_t \cdot \bsdisable{b}^*_{t}, \\
    Q_t &= \hat{v}_{\theta}(\bsdisable{x}_{t}, t, C_{rec}^p, C_{rec}^n) - \hat{v}_{\theta}(\bsdisable{x}_{t}, t, C_{inv}^p, C_{inv}^n), 
  \end{split}
  \label{equ:ode_hat_b_t_approx}
\end{equation}
where $\bsdisable{b}^*_t$ is an approximated $\bsdisable{b}_t$, and $P_t$ is defined in Equ.~\ref{equ:v_theta_approx}.
Equ.~\ref{equ:ode_hat_b_t_approx} is a first-order homogeneous linear differential equation, whose analytical solution is:
\begin{equation}
    \begin{split}
        \bsdisable{b}^*_{t} &= \int_1^t Q_u \exp (\int_u^t P_s ds ) du,
    \end{split}
    \label{equ:analytical_form}
\end{equation}
where $Q_u$ is defined in Equ.~\ref{equ:ode_hat_b_t_approx} and $P_s$ is defined in Equ.~\ref{equ:v_theta_approx}.

This closed-form characterization of $\bsdisable{b}^*_t$ provides a principled means of estimating the bypass term. 
Rather than depending on error-prone terminal inversion states, FlowBypass analytically derives an offset that directly links inversion and reconstruction at any intermediate time $t$. 
This formulation alleviates accumulated discretization errors, ensuring fidelity in irrelevant regions and alignment with edit prompts, while remaining agnostic to backbone architectures, thereby providing a general and powerful framework for training-free image editing.

\subsection{Implementation of FlowBypass}
\label{sec:flowbypass_implement}
Since Equ.~\ref{equ:int_transfer} and Equ.~\ref{equ:analytical_form} involve integral calculations, Euler discretization is employed in the inversion, bypass calculation, and reconstruction to obtain numerical solutions.
Following Stable Diffusion 3.5~\citep{esser2024scaling} and FLUX.1-dev~\citep{batifol2025flux}, we discretize the continuous time interval $[0, 1]$ into a series of $N + 1$ timesteps $\{t_0, t_1, ..., t_N\}$, such that $t_i = \frac{\sigma i}{N + (\sigma - 1) i}$ with shift factor $\sigma=3$.

\textbf{Inversion.}
We first invert the origin image $\bsdisable{x}_{t_0}$ with null prompt $\varnothing$, preserving sufficient structural information for subsequent reconstruction. 
The discretized inversion trajectory is obtained by:
\begin{equation}
    \bsdisable{x}_{t_{i+1}} = \bsdisable{x}_{t_{i}} + (t_{i+1} - t_i) \cdot \hat{v}_{\theta}(\bsdisable{x}_{t_{i}}, t_{i}, C_{inv}^p, C_{inv}^n),
\end{equation}
{where the factor $(t_{i+1}-t_i)>0$ pushes the state towards noise.}
Meanwhile, the terms $\hat{v}_{\theta}(\bsdisable{x}_{t_i}, t_i, C_{rec}^p, C_{rec}^n)$ and $\frac{\partial}{\partial \bsdisable{x}_{t_i}}v_{\theta}(\bsdisable{x}_{t_i}, t_i, C_{rec}^p, C_{rec}^n)$ are also calculated, preparing the calculation of bypass $\bsdisable{b}^*_t$.
Specifically, we set $C_{rec}^p=C_y$ (\textit{i.e.}, target edit prompt) and $C_{rec}^n=C_x$ (\textit{i.e.}, origin prompt) in the reconstruction trajectory.
Importantly, our prompt choice is not an empirical hyperparameter decision but follows directly from the theoretical derivation of the bypass. The detail explanation will be given in Sec.~\ref{sec:ablation_prompt}.

The precise calculation of partial derivative $\frac{\partial}{\partial \bsdisable{x}_{t_i}}\hat{v}_{\theta}$ is computationally prohibitive for large-scale backbones. 
We approximate it via finite differences:
\begin{equation}
    \frac{\partial}{\partial \bsdisable{x}_{t_i}} \hat{v}_{\theta}(\bsdisable{x}_{t_i}, t_i, C_y, C_x) \approx \frac{\hat{v}_{\theta}(\bsdisable{x}_{t_i} + \zeta, t_i, C_y, C_x) - \hat{v}_{\theta}(\bsdisable{x}_{t_i}, t_i, C_y, C_x)}{\zeta},
    \label{equ:approx_v}
\end{equation}
where $\zeta$ is a small positive offset. 
This lightweight approximation introduces negligible bias while avoiding prohibitive memory and compute costs.

\begin{table*}[t]
    \centering
    \caption{Comparison with state-of-the-art image editing methods. OR denotes Optimization-Required, FM denotes Feature-Manipulation, SD is Stable Diffusion, LCM v7 is LCM Dreamshaper v7. \textcolor{red}{Red}, \textcolor{green}{green}, and \textcolor{blue}{blue} highlight the best, second-best, and third-best results, respectively. {The * marks indicate that the official code lacks real-image editing implementations, so we implement them ourselves following the authors’ guidance.}}
    \begin{tabular}{c|c|c|c|c|c|c}
        \toprule
        Method & Backbone & OR? & FM? & LPIPS$\downarrow$ & I.Sim.$\uparrow$ & T.Sim.$\uparrow$ \\
        \midrule
        {P2P*}~\citep{Hertz2023PrompttoPromptIE} & SD1.4 & No & Yes & 0.4990 & 81.04 & \textcolor{green}{26.93} \\
        NTI~\citep{mokady2023null} & SD1.4 & Yes & Yes & 0.5798 & 73.96 & 26.38 \\
        DDCM~\citep{xu2024inversion} & LCM v7 & No & Yes & 0.4507 & 87.14 & \textcolor{blue}{26.62} \\
        IP2P~\citep{brooks2023instructpix2pix} & SD1.4 & Yes & No & 0.6103 & 84.85 & 23.95 \\
        Omni-Gen~\citep{xiao2025omnigen} & Phi-3 & Yes & No & \textcolor{blue}{0.3573} & \textcolor{blue}{87.48} & 25.58 \\
        LEDITS++~\citep{brack2024ledits} & SD 1.5 & No & Yes & \textcolor{green}{0.3554} & 81.54 & 21.73 \\
        RF-Solver~\citep{wang2024taming} & FLUX.1-dev & No & Yes & 0.3880 & 87.32 & 25.30 \\
        RF-Inversion~\citep{rout2024semantic} & FLUX.1-dev & No & No & 0.5659 & 83.35 & 25.71 \\
        {FluxSpace*}~\citep{dalva2025fluxspace} & FLUX.1-dev & No & Yes & 0.8058 & 79.74 & 22.74 \\
        {FireFlow}~\citep{deng2024fireflowfastinversionrectified} & FLUX.1-dev & No & Yes & 0.3850 & 87.01 & 25.69 \\
        {FlowEdit}~\citep{kulikov2025flowedit} & FLUX.1-dev & No & No & 0.3921 & \textcolor{green}{87.90} & 25.28 \\
        \midrule
        FlowBypass & SD3.5 Medium & No & No & 0.4228 & 85.96 & 26.45 \\
        FlowBypass & SD3.5 Large & No & No & 0.4507 & 84.73 & \textcolor{red}{27.09} \\
        FlowBypass & FLUX.1-dev & No & No & \textcolor{red}{0.3425} & \textcolor{red}{88.06} & 25.65 \\
        \bottomrule
    \end{tabular}
    \label{tab:compare_sota}
\end{table*}

\textbf{Bypass Calculation.}
We compute the bypass $\bsdisable{b}_t$ using the trapezoidal variant of Euler discretization applied to Equ.~\ref{equ:analytical_form}:
\begin{equation}
  \begin{split}
    \bsdisable{b}^*_{t_i} &\approx -\frac{1}{2}\sum_{u=i}^{N-1} (t_{u+1} - t_{u}) \cdot (Q'_u \cdot E'_u + Q'_{u+1} \cdot E'_{u+1}), \\
    Q'_u &= \hat{v}_{\theta}(\bsdisable{x}_{t_u}, t_u, C_y, C_x) - \hat{v}_{\theta}(\bsdisable{x}_{t_u}, t_u, \varnothing, \varnothing), \\
    E'_u &= \Gamma \Big(-\frac{1}{2}\sum_{s=i}^{u-1}(t_{s+1}-t_s) \cdot (P'_s + P'_{s+1})\Big), \\
    P'_s &= \frac{\hat{v}_{\theta}(\bsdisable{x}_{t_s} + \zeta, t_s, C_y, C_x) - \hat{v}_{\theta}(\bsdisable{x}_{t_s}, t_s, C_y, C_x)}{\zeta}, \\
    \Gamma(x) &= \begin{cases}
                \exp(x), x \le 0 \\
                x + 1, x > 0
            \end{cases}.
  \end{split}
  \label{equ:discret_form}
\end{equation}

To stabilize numerical evaluation, we replace the exponential with its first-order Taylor approximation in the positive domain, preventing uncontrolled growth.

\textbf{Reconstruction.}
Unlike prior inversion–reconstruction pipelines, FlowBypass constructs the reconstruction trajectory starting from an intermediate state $\bsdisable{y}_{t_B}$, rather than from the inverted noise $\bsdisable{y}_{t_N}$.
The reconstruction trajectory can be described as:
\begin{equation}
    \begin{split}
        \bsdisable{y}_{t_B} &= \bsdisable{x}_{t_B} + \bsdisable{b}_{t_B}, \\
        \bsdisable{y}_{t_{i-1}} &= \bsdisable{y}_{t_{i}} + (t_{i-1} - t_{i}) \cdot \hat{v}_{\theta}(\bsdisable{y}_{t_{i}}, t_i, C_y, C_x), \\
    \end{split}
\end{equation}
where $B$ is a user-specified bypass timestep between $0$ and $N$, {the factor $(t_{i-1}-t_i)<0$ pushes the noisy state to the edit image $y_{t_0}$}. 
A larger $B$ prioritizes alignment with the edit prompt, while a smaller $B$ favors fidelity to the original image. 
The solved $\bsdisable{y}_{t_0}$ corresponds to the desired edited image.

\section{Experiments}
In this section, we present a comprehensive evaluation of our approach. 
Through both qualitative and quantitative analyses, we demonstrate the effectiveness and superiority of our approach. Furthermore, we conduct ablation studies to evaluate the contribution of each component in our design, including backbone robustness, approximation, and prompt choice, while the ablation studies of CFG scale, bypass step, and {hyperparameter $\zeta$} are provided in the Appendix.

\subsection{Experiment Setups}
\label{sec:exp_setup}

\textbf{Baselines, Datasets, and Metrics.}
We compare FlowBypass against diverse state-of-the-art baselines, including DDIM-based approaches such as NTI~\citep{mokady2023null}, DDCM~\citep{xu2024inversion}, IPix2Pix~\citep{brooks2023instructpix2pix}, Omni-Gen~\citep{xiao2025omnigen}, and LEDITS++~\citep{brack2024ledits}, as well as RF-based methods including RF-Solver~\citep{wang2024taming}, RF-Inversion~\citep{rout2024semantic}, FluxSpace~\citep{dalva2025fluxspace}, FireFlow~\citep{deng2024fireflowfastinversionrectified} and FlowEdit~\citep{kulikov2025flowedit}.

For evaluation, we adopt the EditEvalv2 benchmark~\citep{huang2025diffusion}, which contains 150 high-resolution images {across seven sub-tasks: object addition, object replacement, object removal, background change, style change, texture change, and action change}. 
Each image is resized while preserving its aspect ratio, with the longer side scaled to 1024 pixels for computational efficiency.

We assess editing performance from three complementary perspectives: (i) \emph{perceptual fidelity}, measured by LPIPS~\citep{zhang2018unreasonable} between edited and original images; (ii) \emph{semantic fidelity}, quantified by CLIP similarity~\citep{radford2021learning} between edited and original images (\textit{i.e.}, I.Sim.); and (iii) \emph{semantic alignment}, evaluated by CLIPScore~\citep{hessel2021clipscore} between edited images and edit prompts (\textit{i.e.}, T.Sim.).

\begin{figure*}[t]
    \centering
    \includegraphics[width=0.98\linewidth]{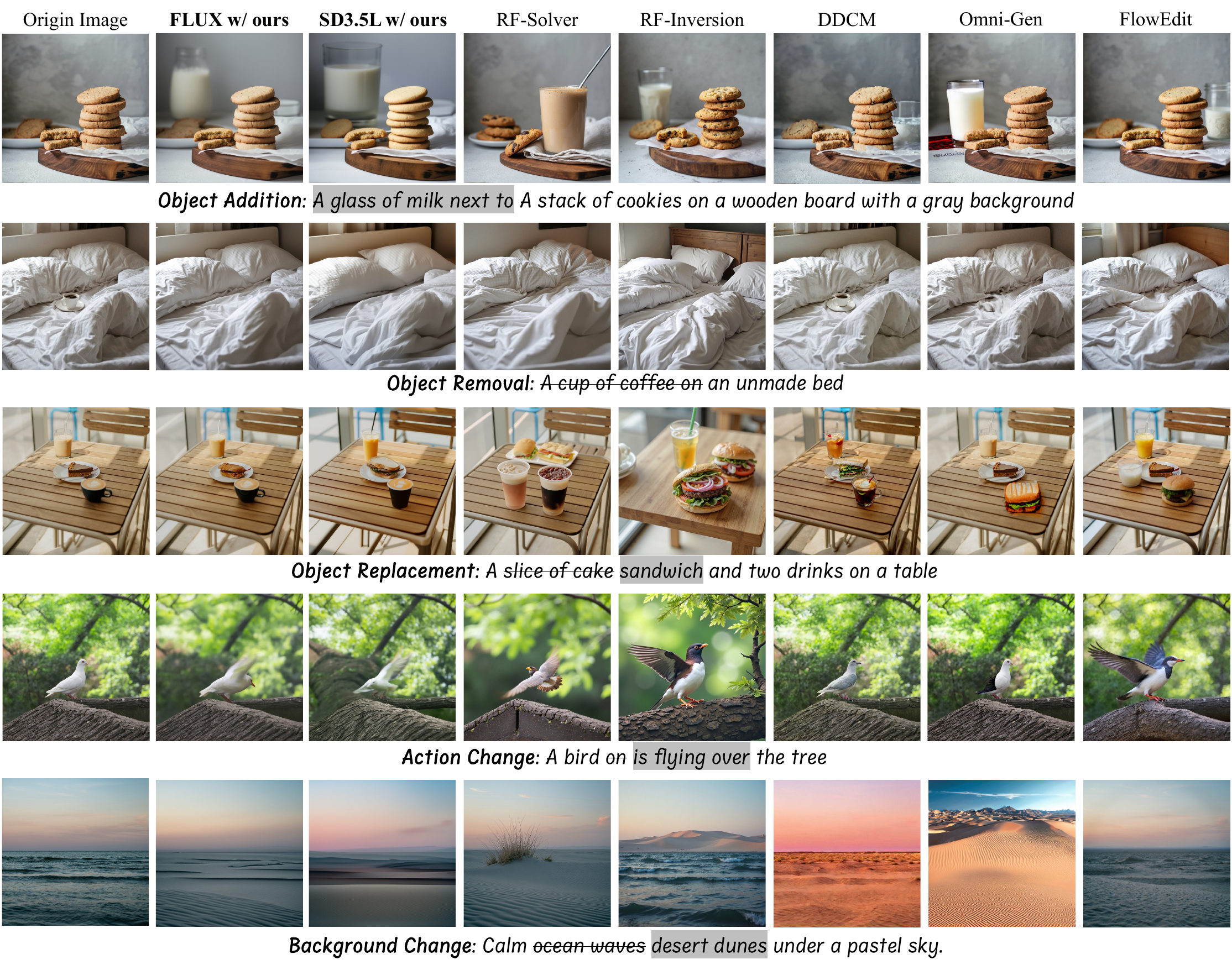}
    \caption{
    \textbf{Qualitative comparison with other image editing methods.} Zoom in for a better view.}
    \label{fig:visual_compare}
\end{figure*}

\textbf{Implementation Details.}
FlowBypass is implemented in PyTorch and all experiments are conducted on a single NVIDIA GeForce RTX 4090 GPU. 
We evaluate on three widely used RF backbones: Stable Diffusion 3.5 Medium (\textit{i.e.}, SD3.5M), Stable Diffusion 3.5 Large (\textit{i.e.}, SD3.5L)~\citep{esser2024scaling}, and FLUX.1-dev (\textit{i.e.}, FLUX)~\citep{batifol2025flux}.

\begin{table}[t]
    \centering
    \small
    \setlength{\tabcolsep}{1.6pt}
    \caption{Performance of different backbones.}
    \begin{tabular}{c|c|c|c|c}
        \toprule
        Backbone & Setting & LPIPS$\downarrow$ & I.Sim.$\uparrow$ & T.Sim.$\uparrow$ \\
        \midrule
        \multirow{3}{*}{SD3.5M} & w/ FlowBypass & 0.4228 & 85.96 & 26.45 \\
        \cmidrule(l{0.2em}r{0.2em}){2-5}
         & Rec. from t=30 & 0.3082 & 90.67 & 24.87 \\
        & Rec. from t=50 & 0.6354 & 77.18 & 27.57 \\
        \midrule
        \multirow{3}{*}{SD3.5L} & w/ FlowBypass & 0.4507 & 84.73 & 27.09 \\
        \cmidrule(l{0.2em}r{0.2em}){2-5}
         & Rec. from t=30 & 0.3288 & 91.01 & 25.36 \\
        & Rec. from t=50 & 0.6576 & 76.95 & 27.94 \\
        \midrule
        \multirow{3}{*}{FLUX} & w/ FlowBypass & 0.3425 & 88.06 & 25.65 \\
        \cmidrule(l{0.2em}r{0.2em}){2-5}
         & Rec. from t=30 & 0.2240 & 94.47 & 23.85 \\
        & Rec. from t=50 & 0.5811 & 78.21 & 27.78 \\
        \bottomrule
    \end{tabular}
    \label{tab:diff_backbone}
\end{table}

\begin{figure}[t]
    \centering
    \includegraphics[width=0.95\linewidth]{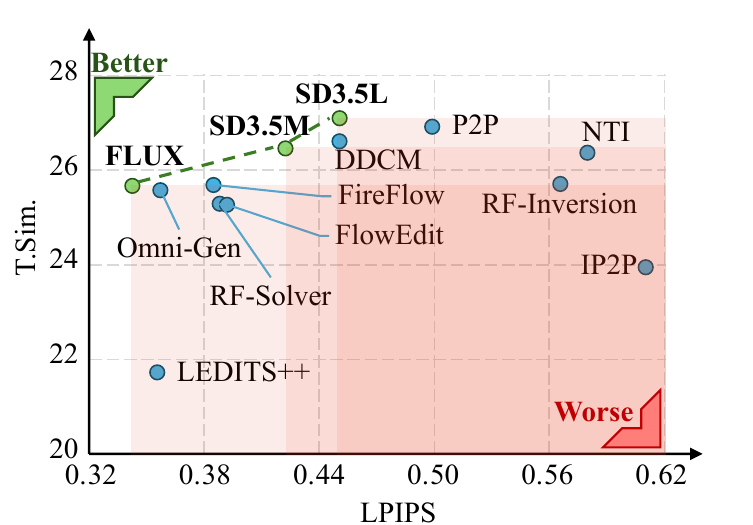}
    \caption{{Scatter figure of quantitative comparison. Green points denote the performance of FlowBypass, whereas blue points denote the performance of other methods. FluxSpace is omitted because its extremely poor performance would compromise the readability of the scatter plot.}}
    \label{fig:compare_sota_scatter}
\end{figure}

\subsection{Quantitative Comparison with State-of-the-Art Methods}
We quantitatively compare FlowBypass with existing editing methods in Tab.~\ref{tab:compare_sota}. 
{The * marks indicate that the official codebase does not provide implementations for real-image editing. Therefore, we implement these components ourselves following the authors’ guidelines.}
Importantly, superior editing performance cannot be judged by fidelity or alignment in isolation; an effective method must achieve a favorable balance across both dimensions.
As reported in Tab.~\ref{tab:compare_sota}, FLUX w/ FlowBypass achieves the highest perceptual fidelity, while SD3.5L w/ FlowBypass demonstrates the best semantic alignment.
More importantly, across all three backbones, FlowBypass consistently delivers strong alignment without substantially compromising fidelity. This trend is clearly reflected in Fig.~\ref{fig:compare_sota_scatter}, where the FlowBypass-related points are positioned closest to the top-left corner, approaching the Pareto frontier of fidelity and alignment.
These results confirm the effectiveness of FlowBypass in achieving superior image editing performance with high fidelity and alignment.

\begin{figure*}[t]
    \centering
    \includegraphics[width=0.9\linewidth]{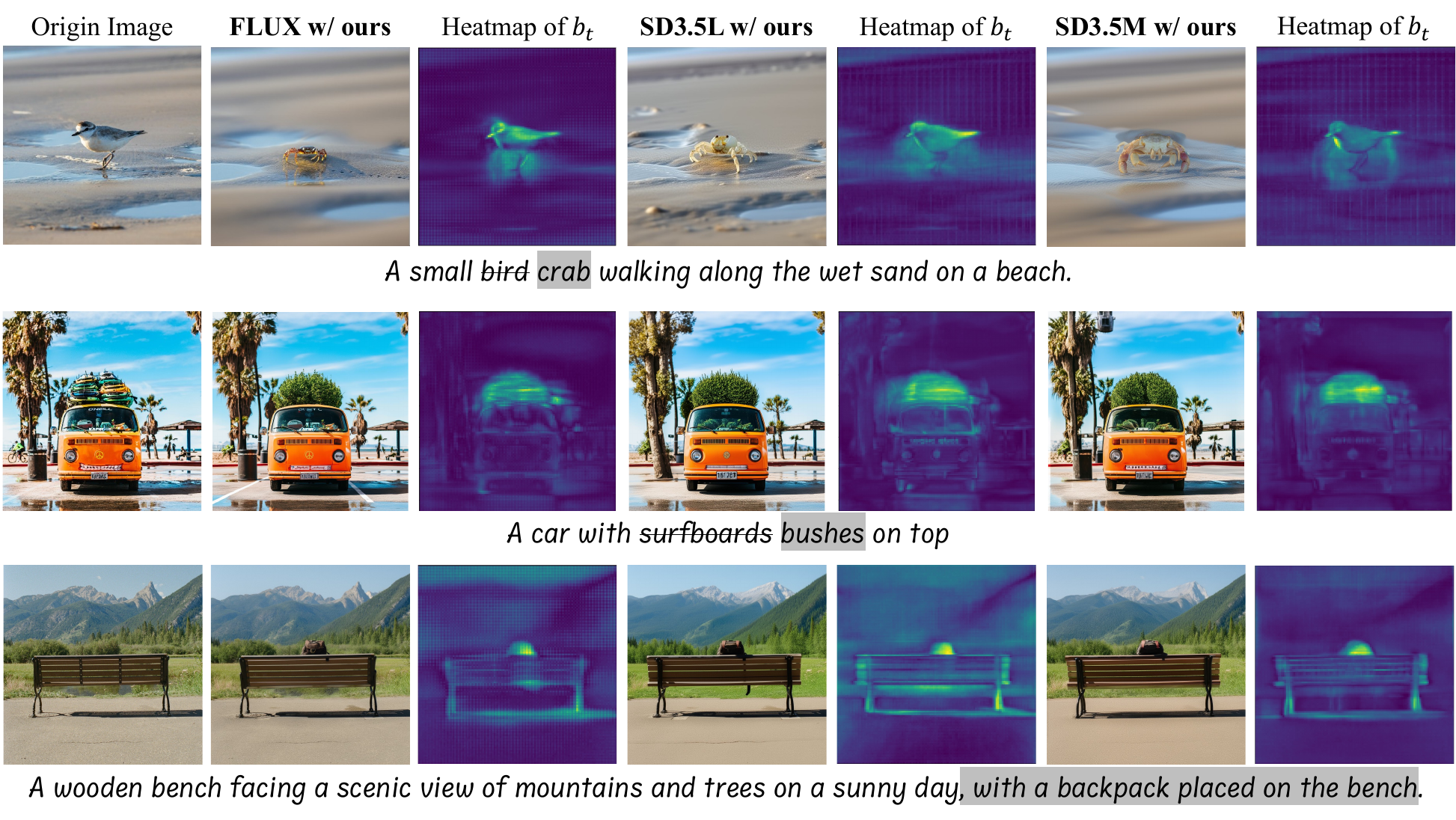}
    \caption{
    \textbf{Visualization of bypass.} Yellow regions indicate higher L1-norm values, while blue regions indicate lower values, reflecting the spatial distribution of bypass magnitude. Zoom in for a better view.}
    \label{fig:bypass_visual}
\end{figure*}

\subsection{Qualitative Comparison with State-of-the-Art Editing Methods}
The qualitative comparison with state-of-the-art image editing baselines is presented in Fig.~\ref{fig:visual_compare}.
To assess the versatility of different approaches, we evaluate their performance across diverse sub-tasks.
Baseline methods often exhibit undesired modifications or incomplete edits, for example, unintentionally altering coffee cups in the Object Replacement sub-task or failing to animate the bird in the Action Change sub-task.
In contrast, FlowBypass consistently delivers edits that closely follow the intended prompts while preserving structural and textural fidelity in regions that should remain untouched.
This balance between alignment and fidelity underscores the robustness of FlowBypass and highlights its superior editing capability.
More qualitative comparison and edited results outside the EditEvalv2 dataset are provided in the Appendix.
{
We also conduct a user study to provide comprehensive subjective evidence of the visual performance of our method in the Appendix.
}

\subsection{Ablation Study}

\subsubsection{Robustness on Different Backbones}
We perform an ablation study to evaluate the robustness of FlowBypass across three representative different backbones, including SD3.5M, SD3.5L, and FLUX.
As mentioned before, the reconstruction trajectory starts from $\bsdisable{y}_{t_B} = \bsdisable{x}_{t_B} + \bsdisable{b}_{t_B}$. 
To verify the alignment contributed by FlowBypass, we conduct two diagnostic settings: (i) we set $\bsdisable{b}_{t_B} = \bsdisable{0}$ and reconstruct from timestep $t=30$, denoted as ``Rec. from $t=30$''; 
and (ii) we execute the entire reconstruction trajectory, starting from $t=50$, to examine the fidelity from FlowBypass, denoted as ``Rec. from $t=50$''.

As shown in Tab.~\ref{tab:diff_backbone}, reconstructing the entire trajectory from $t=50$ degrades fidelity due to error accumulation, whereas disabling the bypass at $t=30$ limits the framework’s ability to perform effective edits in response to the edit prompts. In practice, reconstruction from $t=30$ results in under-edited outputs, while starting from $t=50$ introduces unintended changes, as illustrated by the qualitative results provided in the Appendix.
In contrast, the proposed FlowBypass incorporates an appropriate bypass $\bsdisable{b}_{t_B}$, enabling faithful and accurate editing that achieves a better balance between fidelity and alignment across different backbones.
The visualization in Fig.~\ref{fig:bypass_visual} further confirms that bypasses exhibit strong activations in the regions specified by the edit prompts, while remaining low in irrelevant areas that should stay consistent with the original images.

\subsubsection{Effectiveness of Approximation}
We conduct an ablation study to evaluate the effectiveness of the approximations in Equ.~\ref{equ:approx_v} and Equ.~\ref{equ:discret_form}.
Since precisely computing the gradient is intractable, we set $\frac{\partial}{\partial \bsdisable{x}_{t_i}}\hat{v}_{\theta}$ in Equ.~\ref{equ:approx_v} to zero to assess the contribution of the gradient term in the bypass.
In addition, we perform ablation experiments on the approximation of $\exp$ in Equ.~\ref{equ:discret_form} to examine its role in avoiding numerical instability.
The results summarized in Tab.~\ref{tab:ablation_approximation} indicate that approximating $\frac{\partial}{\partial \bsdisable{x}_{t_i}}\hat{v}_{\theta}$ leads to improved alignment, while introducing the approximation of $\exp$ further enhances both fidelity and alignment.
We observe that neglecting the gradient approximation introduces unrealistic local structures, whereas omitting the approximation of $\exp$ leads to uncontrolled exponential growth, which results in severe artifacts in the edited images. 
More example outputs are provided in the Appendix.

\begin{table}[t]
    \centering
    \small
    \caption{Effectiveness of Approximation.}
    \setlength{\tabcolsep}{1.0pt}
    \begin{tabular}{c|c|c|c|c}
        \toprule
        Backbone & Setting & LPIPS$\downarrow$ & I.Sim.$\uparrow$ & T.Sim.$\uparrow$ \\
        \midrule
        \multirow{5}{*}{SD3.5L} & w/o FlowBypass & 0.3288 & 91.01 & 25.36 \\
        \cmidrule(l{0.2em}r{0.2em}){2-5}
         & w/o Approx. $\frac{\partial}{\partial \bsdisable{x}_{t_i}}\hat{v}_{\theta}$ & 0.4035 & 86.67 & 26.73 \\
        & w/o Approx. $\exp$ & 0.5719 & 78.12 & 26.22 \\
        \cmidrule(l{0.2em}r{0.2em}){2-5}
        & w/ FlowBypass & 0.4507 & 84.73 & 27.09 \\
        \midrule
        \multirow{5}{*}{FLUX} & w/o FlowBypass & 0.2240 & 94.47 & 23.85 \\
        \cmidrule(l{0.2em}r{0.2em}){2-5}
         & w/o Approx. $\frac{\partial}{\partial \bsdisable{x}_{t_i}}\hat{v}_{\theta}$ & 0.3013 & 89.98 & 25.38 \\
        & w/o Approx. $\exp$ & 0.4188 & 85.17 & 25.57 \\
        \cmidrule(l{0.2em}r{0.2em}){2-5}
        & w/ FlowBypass & 0.3425 & 88.06 & 25.65 \\
        \bottomrule
    \end{tabular}    \label{tab:ablation_approximation}
\end{table}

\begin{figure*}[t]
    \centering
    \includegraphics[width=0.9\linewidth]{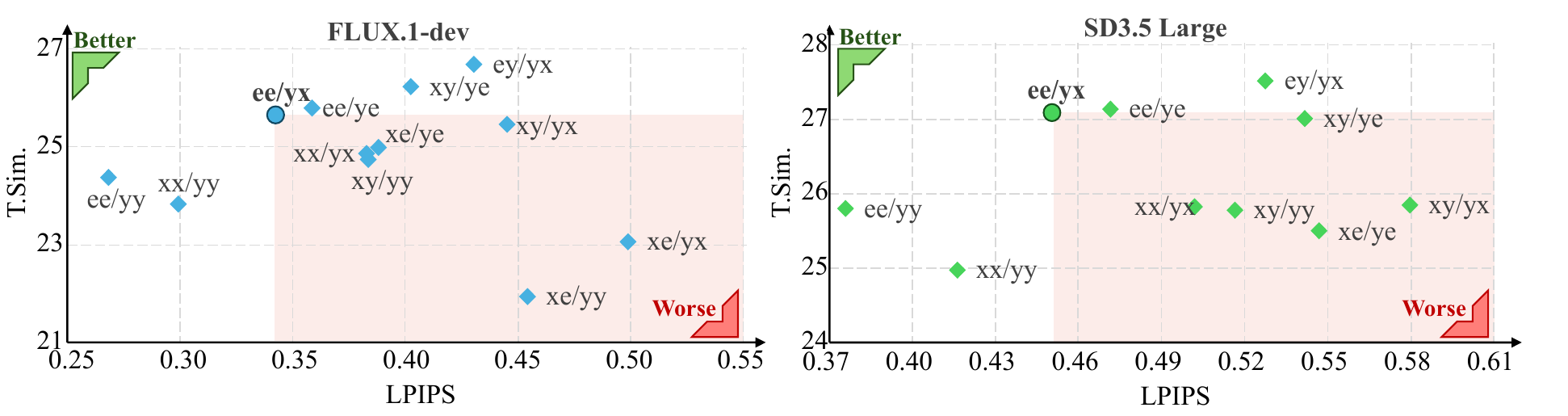}
    \caption{{\textbf{Performance of different prompt choices.} In each notation, the segment before the slash denotes inversion prompts and the segment after the slash denotes reconstruction prompts. Within each segment, the first character indicates the positive prompt and the second indicates the negative prompt, where ``x'' denotes $C_x$, ``y'' denotes $C_y$, and ``e'' denotes $\varnothing$.}}
    \label{fig:prompt_choice}
\end{figure*}

\subsubsection{Impact of Prompt Choice}
\label{sec:ablation_prompt}
The choice of prompts plays a crucial role in FlowBypass.
As shown in Equ.~\ref{equ:analytical_form}, four prompts need to be determined: $C_{inv}^p$, $C_{inv}^n$, $C_{rec}^p$, and $C_{rec}^n$.
To explore this design, we evaluate different reasonable prompt combinations, with results illustrated in Fig.~\ref{fig:prompt_choice}.
The detailed statistical results and qualitative comparison are provided in the Appendix.

Among them, the combination ``ee/yx'' which applies the empty prompt $\varnothing$ during inversion and uses $C_y$ as the positive prompt and $C_x$ as the negative prompt during reconstruction achieves the best balance between fidelity and alignment.
{
We attribute this to two factors: (i) using the empty prompt $\varnothing$ in inversion preserves sufficient structural and semantic information from the origin images in the inversion trajectory, thus benefiting fidelity; and (ii) the balanced non-linear compensation constructed by $C_{rec}^p = C_y, C_{rec}^n = C_x$.
Specifically, in the analytical solution of a first-order linear differential equation, the exponential term acts as the accumulated effect of nonlinear terms along the trajectory. In FlowBypass, this exponential term compensates for the semantic discrepancies introduced by the Taylor expansion. Consequently, using the ``yx'' prompt configuration provides a balanced compensation for this integrating term and amplifies the semantic shift from origin semantics to edit semantics. 
Alternative settings (\textit{e.g.}, ``ye" and ``yy") would create an imbalanced compensation, causing the bypass computation to overly favor either the target or the origin image, thereby harming fidelity or alignment.
}

\section{Conclusion}
In this work, we propose FlowBypass, a general rectified-flow image editing framework that achieves high alignment with target edit prompts while preserving consistency with irrelevant regions of the original images.
It is accomplished by introducing a bypass between the inversion and reconstruction trajectories, without requiring any additional training, test-time optimization, or feature manipulation.
The framework is theoretically motivated by the formulation of bypassing across two trajectories and is realized through carefully designed approximations and prompt selection strategies.
Extensive experiments demonstrate that FlowBypass outperforms existing state-of-the-art methods, striking a superior balance between fidelity and alignment.

\section*{Impact Statement}
This paper presents work whose goal is to advance the field of machine learning. There are many potential societal consequences of our work, none of which we feel must be specifically highlighted here.


\bibliography{example_paper}

@inproceedings{mokady2023null,
  title={Null-text inversion for editing real images using guided diffusion models},
  author={Mokady, Ron and Hertz, Amir and Aberman, Kfir and Pritch, Yael and Cohen-Or, Daniel},
  booktitle={Proceedings of the IEEE/CVF Conference on Computer Vision and Pattern Recognition},
  pages={6038--6047},
  year={2023}
}

@inproceedings{miyake2025negative,
  title={Negative-prompt inversion: Fast image inversion for editing with text-guided diffusion models},
  author={Miyake, Daiki and Iohara, Akihiro and Saito, Yu and Tanaka, Toshiyuki},
  booktitle={2025 IEEE/CVF Winter Conference on Applications of Computer Vision},
  pages={2063--2072},
  year={2025},
  organization={IEEE}
}

@inproceedings{tumanyan2023plug,
  title={Plug-and-play diffusion features for text-driven image-to-image translation},
  author={Tumanyan, Narek and Geyer, Michal and Bagon, Shai and Dekel, Tali},
  booktitle={Proceedings of the IEEE/CVF Conference on Computer Vision and Pattern Recognition},
  pages={1921--1930},
  year={2023}
}

@inproceedings{Cao2023MasaCtrlTM,
  title={MasaCtrl: Tuning-Free Mutual Self-Attention Control for Consistent Image Synthesis and Editing},
  author={Ming Cao and Xintao Wang and Zhongang Qi and Ying Shan and Xiaohu Qie and Yinqiang Zheng},
  booktitle={Proceedings of 2023 IEEE/CVF International Conference on Computer Vision},
  year={2023},
  pages={22503-22513}
}

@inproceedings{xu2024inversion,
  title={Inversion-free image editing with language-guided diffusion models},
  author={Xu, Sihan and Huang, Yidong and Pan, Jiayi and Ma, Ziqiao and Chai, Joyce},
  booktitle={Proceedings of the IEEE/CVF Conference on Computer Vision and Pattern Recognition},
  pages={9452--9461},
  year={2024}
}

@inproceedings{brooks2023instructpix2pix,
  title={Instructpix2pix: Learning to follow image editing instructions},
  author={Brooks, Tim and Holynski, Aleksander and Efros, Alexei A},
  booktitle={Proceedings of the IEEE/CVF Conference on Computer Vision and Pattern Recognition},
  pages={18392--18402},
  year={2023}
}

@inproceedings{brack2024ledits,
  title={Ledits++: Limitless image editing using text-to-image models},
  author={Brack, Manuel and Friedrich, Felix and Kornmeier, Katharia and Tsaban, Linoy and Schramowski, Patrick and Kersting, Kristian and Passos, Apolin{\'a}rio},
  booktitle={Proceedings of the IEEE/CVF Conference on Computer Vision and Pattern Recognition},
  pages={8861--8870},
  year={2024}
}

@inproceedings{xiao2025omnigen,
  title={Omnigen: Unified image generation},
  author={Xiao, Shitao and Wang, Yueze and Zhou, Junjie and Yuan, Huaying and Xing, Xingrun and Yan, Ruiran and Li, Chaofan and Wang, Shuting and Huang, Tiejun and Liu, Zheng},
  booktitle={Proceedings of the Computer Vision and Pattern Recognition Conference},
  pages={13294--13304},
  year={2025}
}

@inproceedings{wang2024taming,
  title={Taming rectified flow for inversion and editing},
  author={Wang, Jiangshan and Pu, Junfu and Qi, Zhongang and Guo, Jiayi and Ma, Yue and Huang, Nisha and Chen, Yuxin and Li, Xiu and Shan, Ying},
  booktitle={Proceedings of the International Conference on Machine Learning},
  year={2025}
}

@inproceedings{rout2024semantic,
  title={Semantic image inversion and editing using rectified stochastic differential equations},
  author={Rout, Litu and Chen, Yujia and Ruiz, Nataniel and Caramanis, Constantine and Shakkottai, Sanjay and Chu, Wen-Sheng},
  booktitle={Proceedings of the International Conference on Learning Representations},
  year={2025}
}

@inproceedings{liu2022flow,
  title={Flow Straight and Fast: Learning to Generate and Transfer Data with Rectified Flow},
  author={Liu, Xingchao and Gong, Chengyue and others},
  booktitle={Proceedings of the International Conference on Learning Representations},
  year={2022}
}

@inproceedings{lipman2023flow,
  title={Flow Matching for Generative Modeling},
  author={Lipman, Yaron and Chen, Ricky TQ and Ben-Hamu, Heli and Nickel, Maximilian and Le, Matthew},
  booktitle={Proceedings of the International Conference on Learning Representations},
  year={2023}
}

@article{huang2025diffusion,
  title={Diffusion model-based image editing: A survey},
  author={Huang, Yi and Huang, Jiancheng and Liu, Yifan and Yan, Mingfu and Lv, Jiaxi and Liu, Jianzhuang and Xiong, Wei and Zhang, He and Cao, Liangliang and Chen, Shifeng},
  journal={IEEE Transactions on Pattern Analysis and Machine Intelligence},
  year={2025},
  publisher={IEEE}
}

@inproceedings{Song2021DenoisingDI,
  title={Denoising Diffusion Implicit Models},
  author={Jiaming Song and Chenlin Meng and Stefano Ermon},
  booktitle={Proceedings of the International Conference on Learning Representations},
  year={2021}
}

@inproceedings{HubermanSpiegelglas2023AnEF,
  title={An Edit Friendly DDPM Noise Space: Inversion and Manipulations},
  author={Inbar Huberman-Spiegelglas and Vladimir B. Kulikov and Tomer Michaeli},
  booktitle={Proceedings of the IEEE/CVF Conference on Computer Vision and Pattern Recognition},
  year={2024},
  pages={12469-12478}
}

@article{Ravi2023PRedItORTG,
  title={PRedItOR: Text Guided Image Editing with Diffusion Prior},
  author={Hareesh Ravi and Sachin Kelkar and Midhun Harikumar and Ajinkya Kale},
  journal={ArXiv},
  year={2023},
  volume={abs/2302.07979}
}

@article{Wang2023InstructEditIA,
  title={InstructEdit: Improving Automatic Masks for Diffusion-based Image Editing With User Instructions},
  author={Qian Wang and Biao Zhang and Michael Birsak and Peter Wonka},
  journal={arXiv preprint arXiv:2305.18047},
  year={2023}
}

@inproceedings{Hertz2023PrompttoPromptIE,
  title={Prompt-to-Prompt Image Editing with Cross Attention Control},
  author={Amir Hertz and Ron Mokady and Jay M. Tenenbaum and Kfir Aberman and Yael Pritch and Daniel Cohen-Or},
  booktitle={Proceedings of the International Conference on Learning Representations},
  year={2023}
}

@inproceedings{simsar2025lime,
  title={Lime: localized image editing via attention regularization in diffusion models},
  author={Simsar, Enis and Tonioni, Alessio and Xian, Yongqin and Hofmann, Thomas and Tombari, Federico},
  booktitle={2025 IEEE/CVF Winter Conference on Applications of Computer Vision},
  pages={222--231},
  year={2025},
  organization={IEEE}
}

@inproceedings{Timofte2017NTIRE2C,
  title={NTIRE 2017 Challenge on Single Image Super-Resolution: Methods and Results},
  author={Radu Timofte and Eirikur Agustsson},
  booktitle={Proceedings of IEEE Conference on Computer Vision and Pattern Recognition Workshops},
  year={2017},
  pages={1110-1121},
}

@inproceedings{feng2025dit4edit,
  title={Dit4edit: Diffusion transformer for image editing},
  author={Feng, Kunyu and Ma, Yue and Wang, Bingyuan and Qi, Chenyang and Chen, Haozhe and Chen, Qifeng and Wang, Zeyu},
  booktitle={Proceedings of the AAAI Conference on Artificial Intelligence},
  volume={39},
  pages={2969--2977},
  year={2025}
}

@inproceedings{yu2025anyedit,
  title={Anyedit: Mastering unified high-quality image editing for any idea},
  author={Yu, Qifan and Chow, Wei and Yue, Zhongqi and Pan, Kaihang and Wu, Yang and Wan, Xiaoyang and Li, Juncheng and Tang, Siliang and Zhang, Hanwang and Zhuang, Yueting},
  booktitle={Proceedings of the Computer Vision and Pattern Recognition Conference},
  pages={26125--26135},
  year={2025}
}

@inproceedings{chen2025unireal,
  title={Unireal: Universal image generation and editing via learning real-world dynamics},
  author={Chen, Xi and Zhang, Zhifei and Zhang, He and Zhou, Yuqian and Kim, Soo Ye and Liu, Qing and Li, Yijun and Zhang, Jianming and Zhao, Nanxuan and Wang, Yilin and others},
  booktitle={Proceedings of the Computer Vision and Pattern Recognition Conference},
  pages={12501--12511},
  year={2025}
}

@article{batifol2025flux,
  title={FLUX. 1 Kontext: Flow Matching for In-Context Image Generation and Editing in Latent Space},
  author={Batifol, Stephen and Blattmann, Andreas and Boesel, Frederic and Consul, Saksham and Diagne, Cyril and Dockhorn, Tim and English, Jack and English, Zion and Esser, Patrick and Kulal, Sumith and others},
  journal={arXiv preprint arXiv:2506.15742},
  year={2025}
}

@inproceedings{esser2024scaling,
  title={Scaling rectified flow transformers for high-resolution image synthesis},
  author={Esser, Patrick and Kulal, Sumith and Blattmann, Andreas and Entezari, Rahim and M{\"u}ller, Jonas and Saini, Harry and Levi, Yam and Lorenz, Dominik and Sauer, Axel and Boesel, Frederic and others},
  booktitle={Proceedings of Forty-first International Conference on Machine Learning},
  year={2024}
}

@inproceedings{zhang2018unreasonable,
  title={The unreasonable effectiveness of deep features as a perceptual metric},
  author={Zhang, Richard and Isola, Phillip and Efros, Alexei A and Shechtman, Eli and Wang, Oliver},
  booktitle={Proceedings of the IEEE Conference on Computer Vision and Pattern Recognition},
  pages={586--595},
  year={2018}
}

@article{hessel2021clipscore,
  title={Clipscore: A reference-free evaluation metric for image captioning},
  author={Hessel, Jack and Holtzman, Ari and Forbes, Maxwell and Bras, Ronan Le and Choi, Yejin},
  journal={arXiv preprint arXiv:2104.08718},
  year={2021}
}

@inproceedings{radford2021learning,
  title={Learning transferable visual models from natural language supervision},
  author={Radford, Alec and Kim, Jong Wook and Hallacy, Chris and Ramesh, Aditya and Goh, Gabriel and Agarwal, Sandhini and Sastry, Girish and Askell, Amanda and Mishkin, Pamela and Clark, Jack and others},
  booktitle={Proceedings of International Conference on Machine Learning},
  pages={8748--8763},
  year={2021}
}

@inproceedings{dalva2025fluxspace,
  title={FluxSpace: Disentangled Semantic Editing in Rectified Flow Models},
  author={Dalva, Yusuf and Venkatesh, Kavana and Yanardag, Pinar},
  booktitle={Proceedings of the Computer Vision and Pattern Recognition Conference},
  pages={13083--13092},
  year={2025}
}

@inproceedings{deng2024fireflowfastinversionrectified,
  title={FireFlow: Fast Inversion of Rectified Flow for Image Semantic Editing}, 
  author={Yingying Deng and Xiangyu He and Changwang Mei and Peisong Wang and Fan Tang},
  booktitle={Proceedings of International Conference on Machine Learning},
  year={2025}
}

@inproceedings{kulikov2025flowedit,
  title={Flowedit: Inversion-free text-based editing using pre-trained flow models},
  author={Kulikov, Vladimir and Kleiner, Matan and Huberman-Spiegelglas, Inbar and Michaeli, Tomer},
  booktitle={Proceedings of the IEEE/CVF International Conference on Computer Vision},
  pages={19721--19730},
  year={2025}
}

@inproceedings{rajbhandari2020zero,
  title={Zero: Memory optimizations toward training trillion parameter models},
  author={Rajbhandari, Samyam and Rasley, Jeff and Ruwase, Olatunji and He, Yuxiong},
  booktitle={SC20: International Conference for High Performance Computing, Networking, Storage and Analysis},
  pages={1--16},
  year={2020},
  organization={IEEE}
}
\bibliographystyle{icml2026}

\newpage
\appendix
\onecolumn

{\huge \textbf{Appendix}}

\section{Solving of ODE}
The ODE described in Equ.~\ref{equ:ode_hat_b_t_approx} can be solved using the Integrating Factor Method.
Equ.~\ref{equ:ode_hat_b_t_approx} can be rewritten as:
\begin{equation}
    \begin{split}
        \frac{d}{dt} \bsdisable{b}^*_t &= Q(t) + P(t) \cdot \bsdisable{b}^*_t, \\
        Q(t) &= \hat{v}_{\theta}(\bsdisable{x}_{t}, t, C_{rec}^p, C_{rec}^n) - \hat{v}_{\theta}(\bsdisable{x}_{t}, t, C_{inv}^p, C_{inv}^n), \\
        P(t) &= \frac{\partial}{\partial \bsdisable{x}_{t}}\hat{v}_{\theta}(\bsdisable{x}_{t}, t, C_{rec}^p, C_{rec}^n),\\
        \bsdisable{b}^*_1 &= \bsdisable{0}.
    \end{split}
\end{equation}

Let 
\begin{equation}
    \begin{split}
        \mu(t) &= \exp \Big(-\int_1^t P(s) ds \Big),
    \end{split}
\end{equation}
and it can be obtained that 
\begin{equation}
    \begin{split}
        \frac{d}{dt}[\mu(t)\bsdisable{b}^*_t] = \mu(t) \frac{d}{dt}\bsdisable{b}^*_t - \mu(t)P(t)\bsdisable{b}^*_t = \mu(t)Q(t).
    \end{split}
\end{equation}

Integrating both sides of the equation with respect to $t$ from 1 yields:
\begin{equation}
    \begin{split}
        \mu(t)\bsdisable{b}^*_t - \mu(1)\bsdisable{b}^*_1 = \int_1^t \mu(u)Q(u)du.
    \end{split}
    \label{equ:appendix_int_result}
\end{equation}

Due to $\bsdisable{b}^*_1 = \bsdisable{0}$ and $\mu(t) > 0$ for any $t$, Equ.~\ref{equ:appendix_int_result} can be rearranged as:
\begin{equation}
    \begin{split}
        \bsdisable{b}^*_t &= \frac{\int_1^t \mu(u)Q(u)du + \bsdisable{0}}{\mu(t)} \\
        &= \exp \Big(\int_1^tP(s)ds \Big) \int_1^t \exp \Big(-\int_1^u P(s)ds\Big) Q(u) du \\
        & = \int_1^t \exp \Big(\int_1^tP(s)ds - \int_1^u P(s)ds\Big) Q(u) du \\
        &= \int_1^t \exp \Big(\int_u^tP(s)ds\Big) Q(u) du \\
        &= \int_1^t \Big(\hat{v}_{\theta}(\bsdisable{x}_{u}, u, C_{rec}^p, C_{rec}^n) - \hat{v}_{\theta}(\bsdisable{x}_{u}, u, C_{inv}^p, C_{inv}^n)\Big) \exp \Big(\int_u^t \frac{\partial}{\partial \bsdisable{x}_s}\hat{v}_{\theta}(\bsdisable{x}_{s}, s, C_{rec}^p, C_{rec}^n) ds \Big) du,
    \end{split}
\end{equation}
which is equal to Equ.~\ref{equ:analytical_form}.

\begin{table}[t]
    \centering
    \caption{{Impact of Prompt Choice.}}
    \begin{tabular}{c|c|c|c|c|c|c|c}
        \toprule
        \multicolumn{2}{c|}{Backbone} & \multicolumn{3}{c|}{FLUX.1-dev} & \multicolumn{3}{c}{SD3.5 Large} \\
        \cmidrule(r{0.2em}){1-2}
        \cmidrule(l{0.2em}r{0.2em}){3-5}
        \cmidrule(l{0.2em}){6-8}
        Inverse & Edit & LPIPS$\downarrow$ & I.Sim.$\uparrow$ & T.Sim.$\uparrow$ & LPIPS$\downarrow$ & I.Sim.$\uparrow$ & T.Sim.$\uparrow$ \\
        \midrule
        (+$C_x$, -$C_y$) & (+$C_y$, -$C_x$) & 0.4454 & 83.41 & 25.47 & 0.5798 & 77.58 & 25.85 \\
        (+$C_x$, -$C_y$) & (+$C_y$, -$\varnothing$) & 0.4024 & 86.07 & 26.23 & 0.5418 & 81.25 & 27.01 \\
        (+$C_x$, -$C_y$) & (+$C_y$, -$C_y$) & 0.3836 & 86.54 & 24.74 & 0.5166 & 81.04 & 25.78 \\
        (+$C_x$, -$\varnothing$) & (+$C_y$, -$C_x$) & 0.4990 & 79.11 & 23.07 & 0.7083 & 68.29 & 20.01 \\
        (+$C_x$, -$\varnothing$) & (+$C_y$, -$\varnothing$) & 0.3881 & 86.19 & 24.99 & 0.5471 & 79.72 & 25.50 \\
        (+$C_x$, -$\varnothing$) & (+$C_y$, -$C_y$) & 0.4542 & 82.28 & 21.95 & 0.6834 & 69.48 & 18.79 \\
        (+$C_x$, -$C_x$) & (+$C_y$, -$C_y$) & 0.2992 & 90.85 & 23.84 & 0.4163 & 87.15 & 24.98 \\
        (+$C_x$, -$C_x$) & (+$C_y$, -$C_x$) & 0.3828 & 86.10 & 24.87 & 0.5022 & 81.40 & 25.83 \\
        (+$\varnothing$, -$C_y$) & (+$C_y$, -$C_x$) & 0.4305 & 85.66 & 26.69 & 0.5277 & 82.46 & 27.52 \\
        {(+$\varnothing$, -$\varnothing$)} & {(+$C_y$, -$C_y$)} & 0.2685 & 93.16 & 24.38 & 0.3759 & 90.00 & 25.80  \\
        {(+$\varnothing$, -$\varnothing$)} & {(+$C_y$, -$\varnothing$)} & 0.3586 & 89.65 & 25.80 & 0.4716 & 85.55 & 27.14  \\
        (+$\varnothing$, -$\varnothing$) & (+$C_y$, -$C_x$) & \textbf{0.3425} & \textbf{88.06} & \textbf{25.65} & \textbf{0.4507} & \textbf{84.73} & \textbf{27.09} \\
        \bottomrule
    \end{tabular}    \label{tab:ablation_prompt_choice}
\end{table}

\begin{figure}
    \centering
    \includegraphics[width=1.0\linewidth]{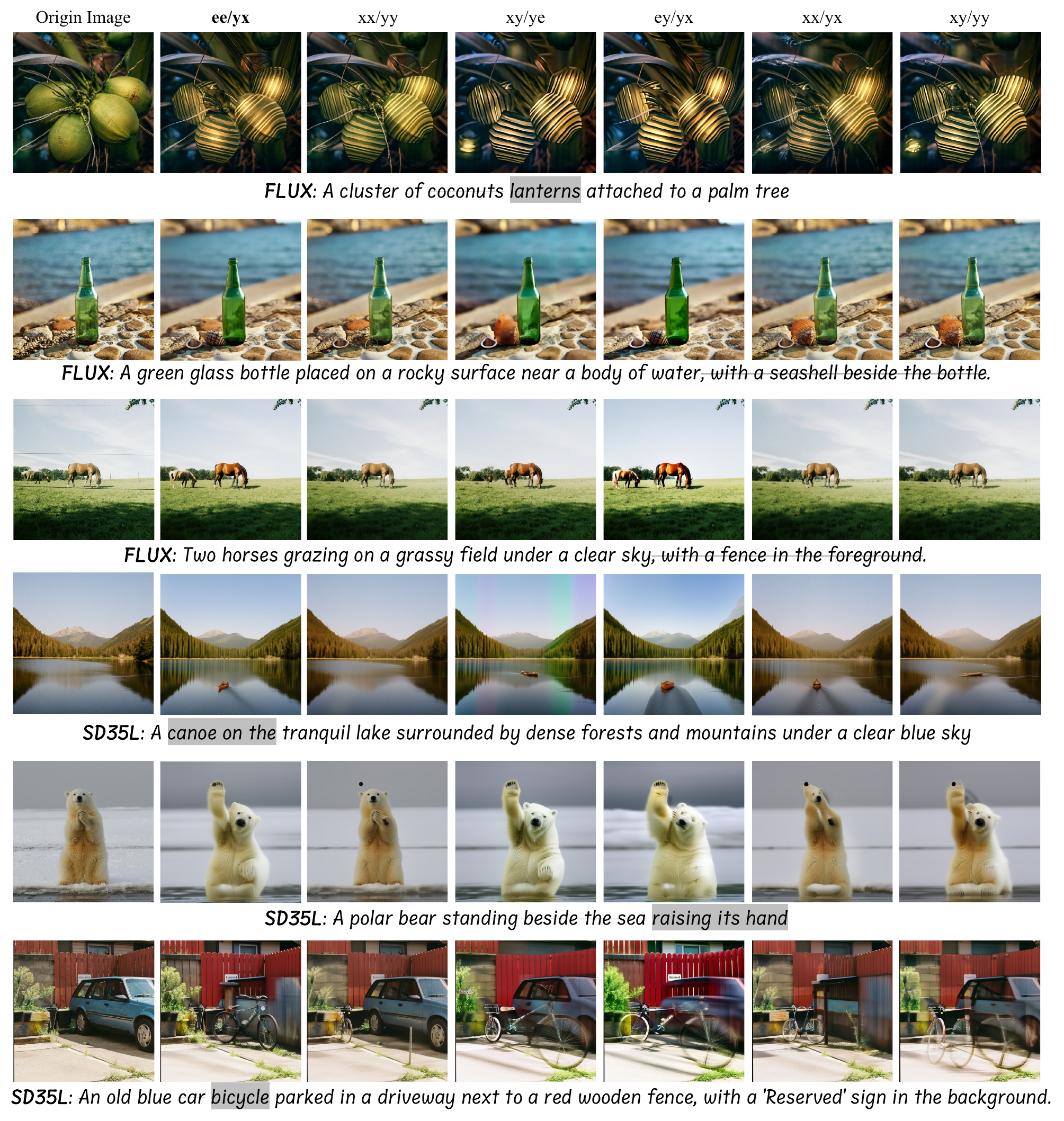}
    \caption{
    \textbf{Visual comparison of ablation results on different prompt choices.} Zoom in for a better view.}
    \label{fig:prompt_choice_visual}
\end{figure} 

\begin{figure}[t]
    \centering
    \includegraphics[width=1.0\linewidth]{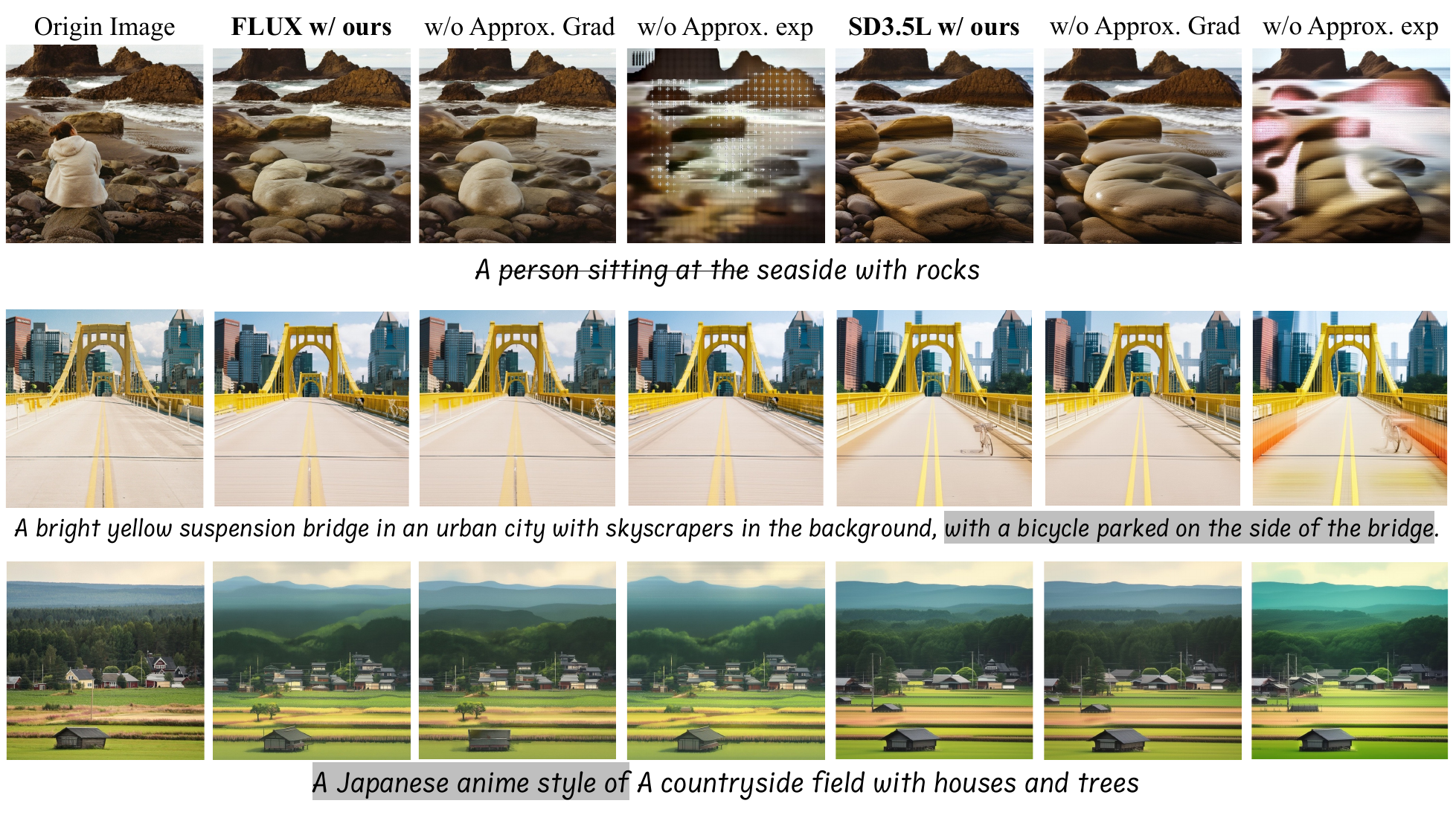}
    \caption{
    \textbf{Visual comparison of ablation results on approximation.} Zoom in for a better view.}
    \label{fig:ablation_approx_figure}
\end{figure}

\section{Additional Experiment Results}

\subsection{Detailed Statistical Analysis of Ablation on Prompt Choice}
The detailed statistical results of Fig.~\ref{fig:prompt_choice} in main text is demonstrated in Tab.~\ref{tab:ablation_prompt_choice}.
Besides, we provide the visual comparison with different prompt choices in Fig.~\ref{fig:prompt_choice_visual}, which indicates that the combination ``ee/yx'' preserves the fidelity of irrelevant regions while successfully applying the intended edit.

\subsection{Visualization of Approximation}
Fig.~\ref{fig:ablation_approx_figure} shows the ablation results on approximation.
Severe artifacts appear when the approximation of $\exp$ is omitted, as illustrated in the fourth and seventh columns. 
{We argue that these artifacts mainly arise from exponential explosion, which injects excessively large values into the calculated bypass, causing the normalization layers in the denoiser network to behave improperly, and finally introduces these unpleasant artifacts.}
Furthermore, removing the approximation of $\frac{\partial}{\partial \bsdisable{x}_{t_i}}\hat{v}_{\theta}$ introduces unrealistic structural details, such as distorted stones in the first row, deformed bicycles in the second row, and abnormal house shapes in the third row.
These visualizations indicate that the introduced approximations not only produce more realistic details but also effectively suppress artifacts that may arise from exponential explosion.

\section{More Experiment Results}

\begin{figure}[t]
    \centering
    \includegraphics[width=\linewidth]{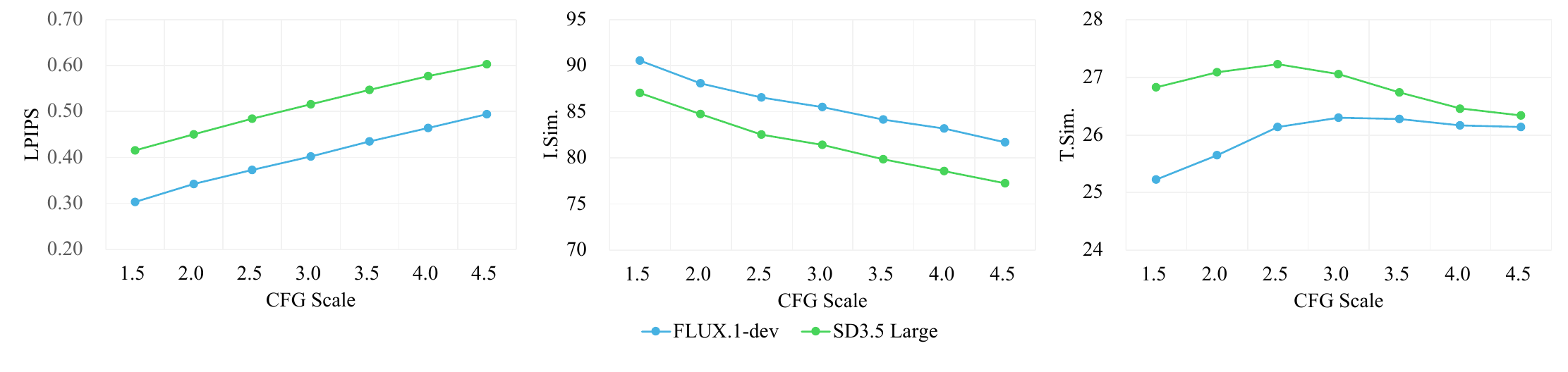}
    \caption{Trends of edit performance as CFG scale $\omega$ increases.}
    \label{fig:cfg_scale}
\end{figure}

\begin{table}[t]
    \centering
    \caption{Impact of CFG scale.}
    \begin{tabular}{c|c|c|c|c|c|c}
        \toprule
        Backbone & \multicolumn{3}{c|}{FLUX.1-dev} & \multicolumn{3}{c}{SD3.5 Large} \\
        \cmidrule(r{0.2em}){1-1}
        \cmidrule(l{0.2em}r{0.2em}){2-4}
        \cmidrule(l{0.2em}){5-7}
        CFG scale & LPIPS$\downarrow$ & I.Sim.$\uparrow$ & T.Sim.$\uparrow$ & LPIPS$\downarrow$ & I.Sim.$\uparrow$ & T.Sim.$\uparrow$ \\
        \midrule
        1.5 & 0.3036 & 90.55 & 25.23 & 0.4155 & 87.04 & 26.83 \\
        2.0 & \textbf{0.3425} & \textbf{88.06} & \textbf{25.65} & \textbf{0.4507} & \textbf{84.73} & \textbf{27.09} \\
        2.5 & 0.3731 & 86.55 & 26.14 & 0.4848 & 82.51 & 27.23 \\
        3.0 & 0.4021 & 85.49 & 26.30 & 0.5160 & 81.40 & 27.06 \\
        3.5 & 0.4350 & 84.13 & 26.28 & 0.5472 & 79.84 & 26.74 \\
        4.0 & 0.4642 & 83.18 & 26.17 & 0.5773 & 78.56 & 26.46 \\
        4.5 & 0.4940 & 81.69 & 26.14 & 0.6027 & 77.24 & 26.34 \\
        \bottomrule
    \end{tabular}
    \label{tab:ablation_cfg_scale}
\end{table}

\begin{table}[t]
    \centering
    \caption{Impact of bypass step.}
    \begin{tabular}{c|c|c|c|c|c|c}
        \toprule
        Backbone & \multicolumn{3}{c|}{FLUX.1-dev} & \multicolumn{3}{c}{SD3.5 Large} \\
        \cmidrule(r{0.2em}){1-1}
        \cmidrule(l{0.2em}r{0.2em}){2-4}
        \cmidrule(l{0.2em}){5-7}
        Timestep & LPIPS$\downarrow$ & I.Sim.$\uparrow$ & T.Sim.$\uparrow$ & LPIPS$\downarrow$ & I.Sim.$\uparrow$ & T.Sim.$\uparrow$ \\
        \midrule
        10 & 0.1849 & 96.14 & 23.36 & 0.3003 & 92.84 & 24.50 \\
        20 & 0.2525 & 92.99 & 24.66 & 0.3695 & 89.37 & 25.86 \\
        30 & \textbf{0.3425} & \textbf{88.06} & \textbf{25.65} & \textbf{0.4507} & \textbf{84.73} & \textbf{27.09} \\
        40 & 0.4490 & 84.25 & 27.16 & 0.5618 & 80.20 & 27.70 \\
        50 & 0.5811 & 78.21 & 27.78 & 0.6576 & 76.95 & 27.94 \\
        \bottomrule
    \end{tabular}
    \label{tab:ablation_bypass_step}
\end{table}

\begin{figure}[t]
    \centering
    \includegraphics[width=\linewidth]{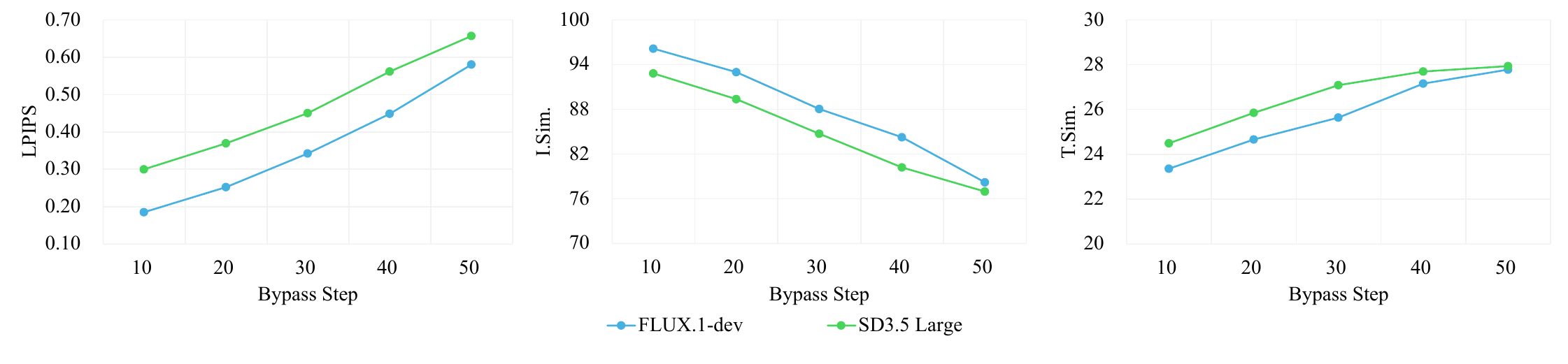}
    \caption{Trends of edit performance as Bypass step $t_B$ changes.}
    \label{fig:bypass_step}
\end{figure}

\subsection{More Qualitative Comparison with SOTA Editing Methods}
\begin{figure}
    \centering
    \includegraphics[width=1.0\linewidth]{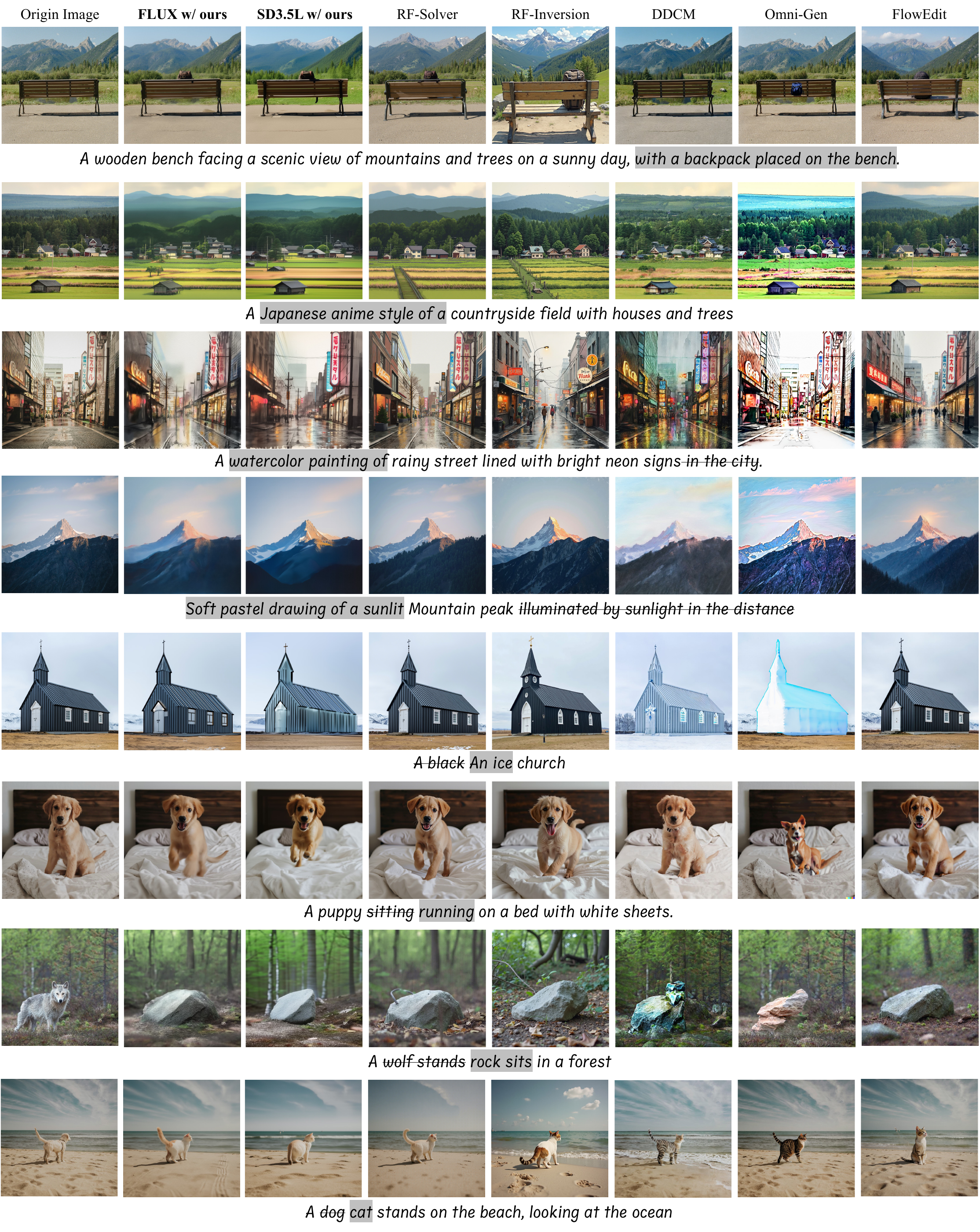}
    \caption{
    {\textbf{More qualitative comparison with SOTA editing methods.} Zoom in for a better view.}}
    \label{fig:visual_compare_supp}
\end{figure}

Additional qualitative comparisons are provided in Fig.~\ref{fig:visual_compare_supp}.
The results demonstrate that FlowBypass, across different backbones, outperforms other editing methods in terms of both fidelity and alignment.

\subsection{{User Study}}
\begin{figure}
    \centering
    \includegraphics[width=1.0\linewidth]{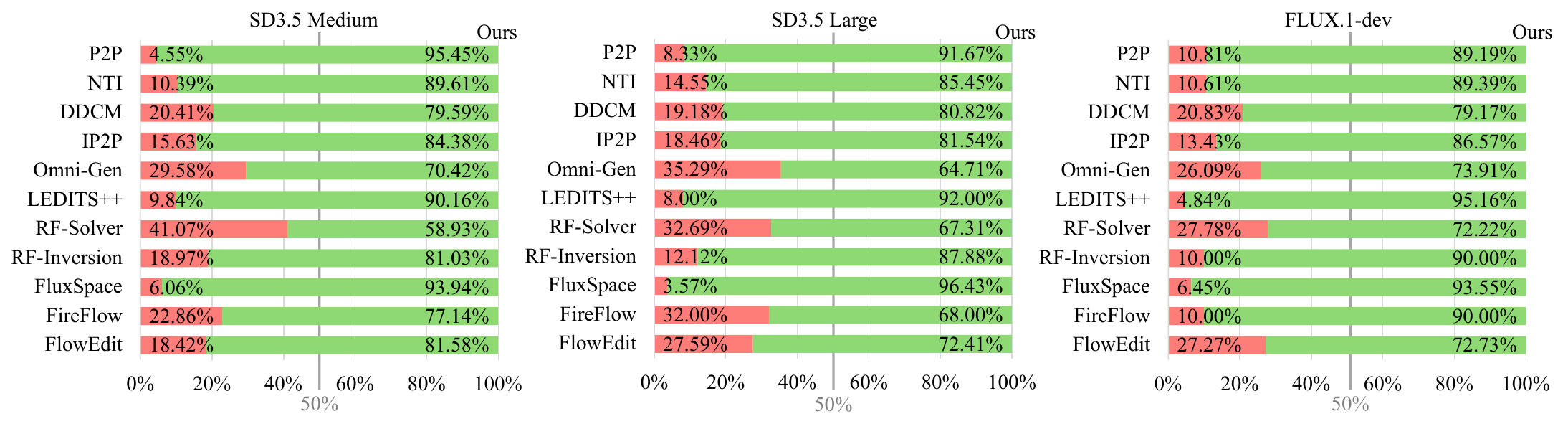}
    \caption{{User Study Statistics. The \textcolor{green}{green} portion represents the percentage of users who rated our method's output as better, while the \textcolor{red}{red} portion represents the percentage who preferred the output of the compared method.}}
    \label{fig:user_study}
\end{figure}

{
A user study is conducted to provide comprehensive subjective evidence of the visual performance of our method.
Participants were asked to compare our results with those produced by state-of-the-art approaches in a pairwise preference setting under controlled viewing conditions.
As shown in Fig.~\ref{fig:user_study}, our method consistently achieves the highest preference rates across all backbones and comparison methods.
In each bar, the green region indicates the proportion of users who preferred our results, while the red region corresponds to the competing method.
These results demonstrate that FlowBypass not only improves objective reconstruction fidelity but also delivers outputs that align more closely with human perception.
}

\subsection{{Runtime Comparison with SOTA Editing Methods}}
{
We present a comprehensive runtime comparison against state-of-the-art editing methods in Tab.~\ref{tab:runtime_compare}. As specified in Sec.~\ref{sec:exp_setup}, all experiments are conducted on a single RTX 4090 GPU. However, FLUX.1-dev cannot be executed naively under the 24GB VRAM limit of the RTX 4090. To ensure functional equivalence while fitting the model into memory, we employ DeepSpeed with ZeRO Stage-3 optimization~\cite{rajbhandari2020zero}, which preserves computational correctness while reducing memory consumption.
It is worth noting that the runtime of FLUX on RTX 4090 is sensitive to system-level factors such as PCIe bandwidth and host RAM capacity. To provide a fairer comparison, we additionally benchmark all methods on a larger-memory GPU (L20) without DeepSpeed, as reported in the fifth column of Tab.~\ref{tab:runtime_compare}.
Although FlowBypass is not the fastest method in absolute runtime, it achieves a practical processing cost for high-resolution image editing while offering superior performance and a well-balanced between fidelity and alignment.
}

{
Additionally, we break down the computational cost into different stages to highlight the extra overhead introduced by the bypass, as shown in Tab.~\ref{tab:runtime_breakdown}.
Theoretically, the bypass computation should account for only about 21.1\% additional cost, and our practical results confirm this, demonstrating that the bypass is not the primary contributor to the overall computational burden.
The discrepancy between the totals reported in Tab.~\ref{tab:runtime_breakdown} and Tab.~\ref{tab:runtime_compare} comes from the exclusion of text encoding time in Tab.~\ref{tab:runtime_breakdown}.
}

\begin{table}[t]
    \centering
    \caption{{Runtime comparison with SOTA editing methods at 1024 $\times$ 1024 resolution. The \# in parentheses indicates the ranking.}}
    \begin{tabular}{c|c|c|c|c}
        \toprule
        Method & Backbone & Precision & 4090 Runtime (s)$\downarrow$ & L20 Runtime (s)$\downarrow$ \\
        \midrule
        P2P* & SD1.4 & FP32 & 17.59 (\#3) & 24.41 (\#4)\\
        NTI & SD1.4 & FP32 & 424.03 (\#14) & 651.65 (\#14)\\
        DDCM & LCM v7 & FP16 & 5.36 (\#1) & 6.47 (\#1) \\
        IP2P & SD1.4 & FP32 & 8.28 (\#2) & 22.16 (\#3)\\
        Omni-Gen & Phi-3 & FP32 & 81.59 (\#8) & 101.84 (\#11)\\
        LEDITS++ & SD 1.5 & FP32 & 33.19 (\#6) & 45.18 (\#7)\\
        RF-Solver & FLUX.1-dev & BF16 & 256.23 (\#13) & 138.33 (\#12)\\
        RF-Inversion & FLUX.1-dev & BF16 & 117.24 (\#10) & 51.78 (\#8)\\
        FluxSpace* & FLUX.1-dev & BF16 & 145.50 (\#11) & 65.75 (\#9)\\
        FireFlow & FLUX.1-dev & BF16 & 29.85 (\#5) & 20.66 (\#2)\\
        FlowEdit & FLUX.1-dev & BF16 & 101.08 (\#9) & 43.95 (\#6)\\
        \midrule
        FlowBypass & SD3.5 Medium & FP16 & 29.43 (\#4) & 35.90 (\#5)\\
        FlowBypass & SD3.5 Large & FP16 & 64.11 (\#7) & 81.31 (\#10)\\
        FlowBypass & FLUX.1-dev & BF16 & 172.24 (\#12) & 192.23 (\#13)\\
        \bottomrule
    \end{tabular}
    \label{tab:runtime_compare}
\end{table}

\begin{table}[t]
    \centering
    \caption{{Computational cost breakdown (second). The \% in parentheses indicates the percentage.}}
    \begin{tabular}{c|c|c|c}
        \toprule
        Stage & SD35M & SD35L & FLUX.1 \\
        \midrule
        VAE encode & 0.23 (1.08\%) & 0.16 (0.31\%) & 0.22 (0.13\%)\\
        Inversion & 9.71 (46.25\%) & 24.38 (46.97\%) & 78.50 (46.52\%)\\
        \textbf{Inversion for Bypass} & 4.32 (20.56\%) & 10.84 (20.87\%) & 34.89 (20.67\%)\\
        \textbf{Bypass} & 0.00836 (0.0398\%) & 0.00335 (0.006461\%) & 0.00359 (0.002129\%)\\
        Recon & 6.42 (30.56\%) & 16.27 (31.34\%) & 54.90 (32.53\%)\\
        VAE decode & 0.32 (1.51\%) & 0.26 (0.50\%) & 0.24 (0.14\%)\\
        \bottomrule
    \end{tabular}
    \label{tab:runtime_breakdown}
\end{table}

\subsection{More Edited Results outside Dataset}
\begin{figure}
    \centering
    \includegraphics[width=1.0\linewidth]{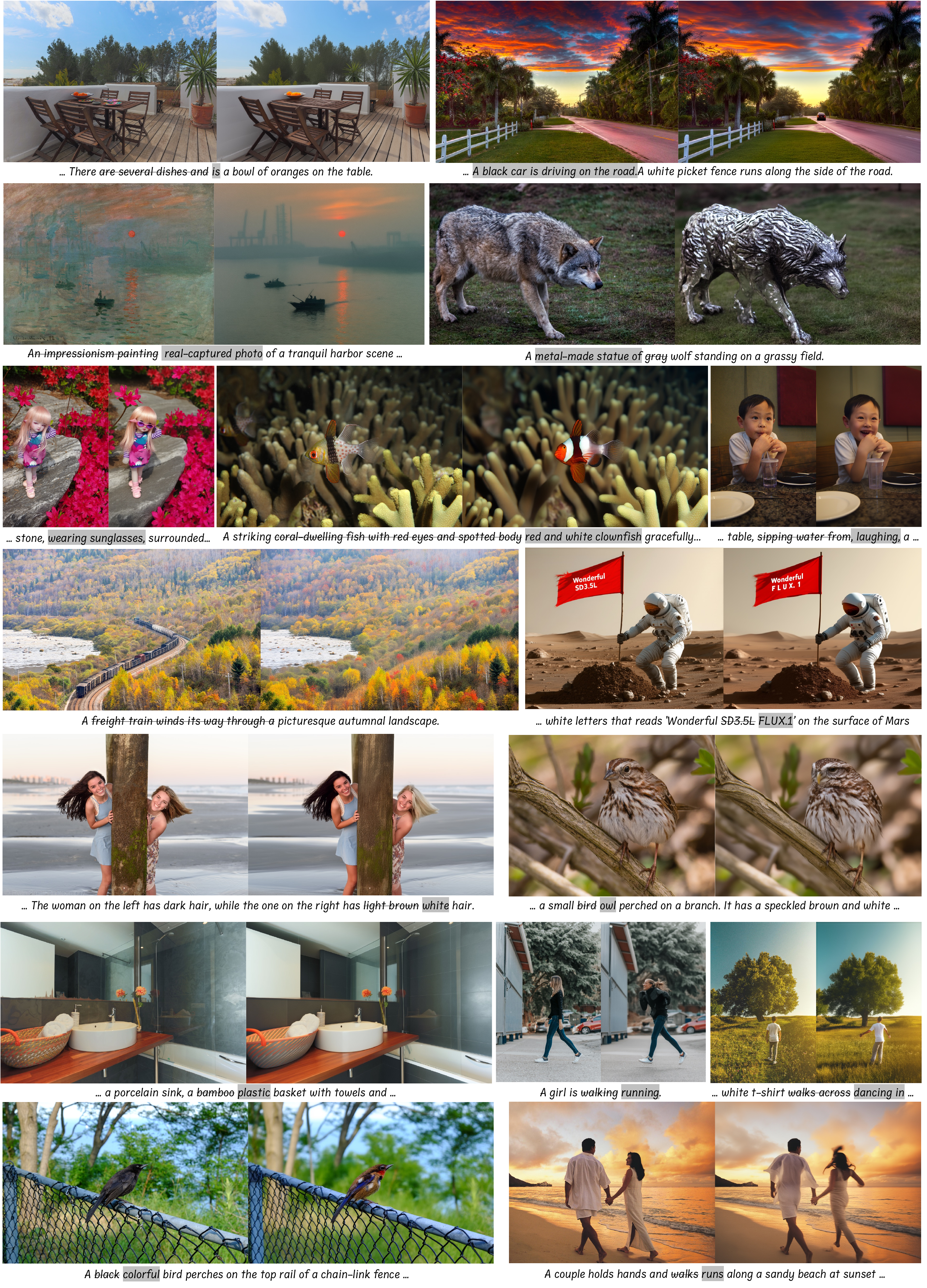}
    \caption{
    \textbf{Edit results outside the dataset.} Zoom in for a better view.}
    \label{fig:outside_result}
\end{figure}

To further evaluate the generalization ability of FlowBypass, we conduct image editing experiments on samples outside the EditEvalv2 dataset, as shown in Fig.~\ref{fig:outside_result}.
{The origin images are selected from four sources, namely the impressionist painting of Claude Monet (first of row 2), the natural image dataset Flickr2K~\citep{Timofte2017NTIRE2C}, the image generated by SD3.5L (second of row 4), and Pexels online website\footnote{[Online] Available at https://www.pexels.com} (second and third of row 6).} 
The results indicate that FlowBypass can effectively edit diverse types of images with fidelity and alignment, which demonstrates its general editing capability.
We also observe that the fidelity is slightly lower when editing the SD3.5L-generated image compared with natural images, as shown in the last of fourth row. This difference may be caused by the domain gap of generation characteristics between FLUX and SD3.5L.
In general, FlowBypass can still handle images that follow the unseen distribution during training to provide outputs with desired editing and fidelity to the original image.

\subsection{Impact of CFG Scale}
\begin{figure}
    \centering
    \includegraphics[width=1.0\linewidth]{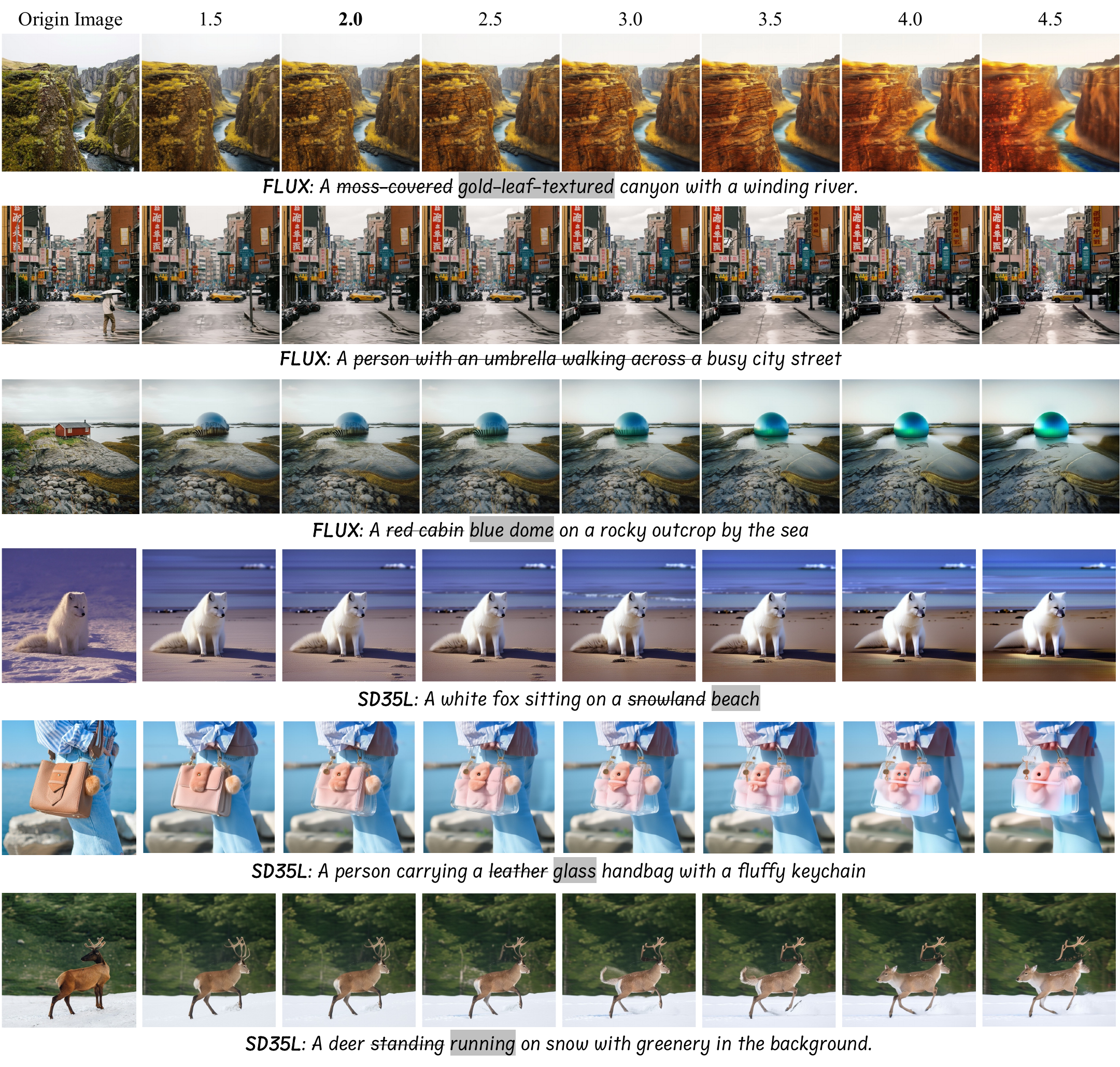}
    \caption{
    \textbf{Visual comparison of ablation results on different CFG scales.} Zoom in for a better view.}
    \label{fig:cfg_scale_visual}
\end{figure}

We conduct ablation experiments on the CFG scale $\omega$ to examine the effect of different guidance strengths on editing performance, with results presented in Fig.~\ref{fig:cfg_scale}.
As the CFG scale increases, fidelity gradually decreases, while alignment first improves and then declines.
When the CFG scale becomes excessively large, overly strong guidance introduces noticeable artifacts and abnormal appearances.
Fig.~\ref{fig:cfg_scale_visual} illustrates this trend, showing that smaller CFG scales preserve fidelity but limit alignment, whereas larger CFG scales behave more aggressively and may even introduce severe artifacts.
Based on these findings, we set the CFG scale $\omega$ to $2$ in our experiments.
The detailed statistical results are provided in Tab.~\ref{tab:ablation_cfg_scale}.

\subsection{{Impact of Hyperparameter $\zeta$}}
{We perform ablation study on the hyperparameter $\zeta$ in Equ.~\ref{equ:approx_v}.
The results in Tab.~\ref{tab:ablation_zeta} indicate that FlowBypass is reasonably robust to $\zeta$, and varying it within a practical range does not meaningfully influence editing quality, which partially validates the assumption underlying the Taylor expansion in Equ.~\ref{equ:approx_v}.}

\begin{table}[t]
    \centering
    \caption{{Impact of hyperparameter $\zeta$.}}
    \begin{tabular}{c|c|c|c|c|c|c}
        \toprule
        Backbone & \multicolumn{3}{c|}{FLUX.1-dev} & \multicolumn{3}{c}{SD3.5 Large} \\
        \cmidrule(r{0.2em}){1-1}
        \cmidrule(l{0.2em}r{0.2em}){2-4}
        \cmidrule(l{0.2em}){5-7}
        $\zeta$ & LPIPS$\downarrow$ & I.Sim.$\uparrow$ & T.Sim.$\uparrow$ & LPIPS$\downarrow$ & I.Sim.$\uparrow$ & T.Sim.$\uparrow$ \\
        \midrule
        0.001 & 0.3290 & 88.56 & 25.66 & 0.4513 & 84.77 & 27.10 \\
        0.005 & 0.3446 & 87.91 & 25.77 & 0.4488 & 85.09 & 27.14 \\
        0.01 & \textbf{0.3425} & \textbf{88.06} & \textbf{25.65} & \textbf{0.4507} & \textbf{84.73} & \textbf{27.09} \\
        0.05 & 0.3509 & 88.03 & 25.84 & 0.4606 & 84.65 & 27.21 \\
        0.1 & 0.3434 & 88.25 & 25.82 & 0.4562 & 84.81 & 27.32 \\
        \bottomrule
    \end{tabular}
    \label{tab:ablation_zeta}
\end{table}

\subsection{Impact of Bypass Step}
\begin{figure}
    \centering
    \includegraphics[width=0.8\linewidth]{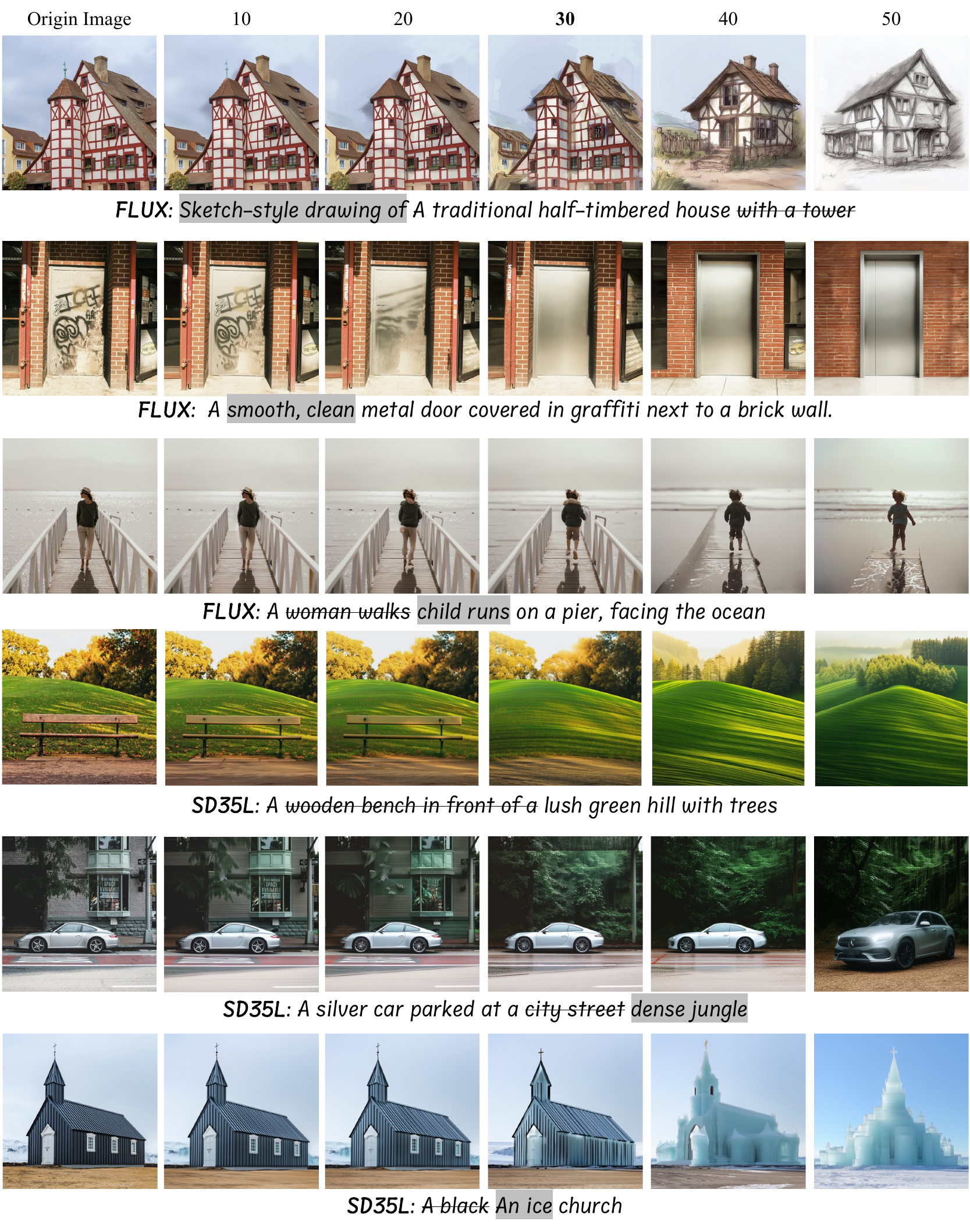}
    \caption{
    \textbf{Visual comparison of ablation results on different bypass steps.} Zoom in for a better view.}
    \label{fig:bypass_step_visual}
\end{figure}

We conduct an ablation study about the impact of bypass step $t_B$, whose trends are presented in Fig.~\ref{fig:bypass_step}.
The results reveal a clear trade-off between fidelity and alignment, where fidelity decreases monotonically and alignment increases monotonically as $t_B$ increases, as demonstrated in Fig.~\ref{fig:bypass_step_visual}.
We choose $t_B=30$ for the best balance between fidelity and alignment.
The detailed statistical results are provided in Tab.~\ref{tab:ablation_bypass_step}.

\subsection{{Ablation Study on Different Reconstruction Timestep without Bypass}}
\begin{figure}
    \centering
    \includegraphics[width=0.6\linewidth]{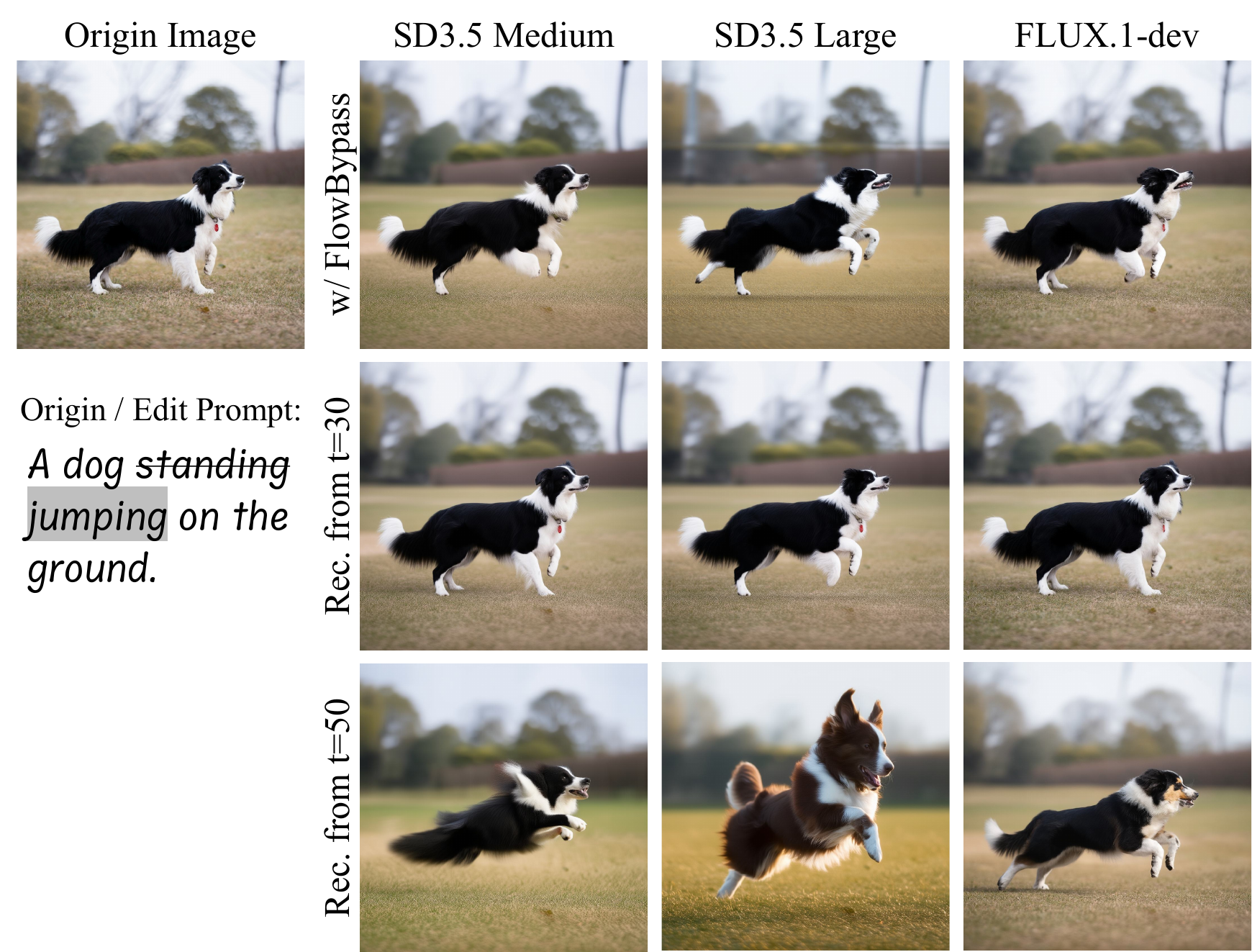}
    \caption{\textbf{Example outputs from different backbones and different reconstruction timesteps.} Zoom in for a better view.}
    \label{fig:diff_backbone_visual}
\end{figure}

{We conduct an ablation study to evaluate the performance of the inversion–reconstruction editing paradigm when no bypass is introduced, and the reconstruction starts not from pure noise but from an intermediate latent state.
As shown in Tab.~\ref{tab:ablation_no_bypass_step}, when starting from larger $t$, fidelity is difficult to maintain, whereas starting from smaller $t$ compromises alignment. 
This observation is further supported by the visual examples in Fig.~\ref{fig:diff_backbone_visual}, where direct reconstruction from $t=30$ yields under-edited outputs, whereas reconstruction from $t=50$ introduces spurious changes, such as altering the dog’s appearance.
This phenomenon provides evidence from opposite that FlowBypass addresses the trade-off by correctly jumping the reconstruction starting point onto the reconstruction trajectory, achieving a balance between fidelity and alignment and yielding more stable, high-quality edits.}

\begin{table}[t]
    \centering
    \caption{{Ablation study on different reconstruction timesteps without bypass.}}
    \begin{tabular}{c|c|c|c|c|c|c}
        \toprule
        Backbone & \multicolumn{3}{c|}{FLUX.1-dev} & \multicolumn{3}{c}{SD3.5 Large} \\
        \cmidrule(r{0.2em}){1-1}
        \cmidrule(l{0.2em}r{0.2em}){2-4}
        \cmidrule(l{0.2em}){5-7}
        Rec. t & LPIPS$\downarrow$ & I.Sim.$\uparrow$ & T.Sim.$\uparrow$ & LPIPS$\downarrow$ & I.Sim.$\uparrow$ & T.Sim.$\uparrow$ \\
        \midrule
        10 & 0.1585 & 97.56 & 22.35 & 0.2561 & 95.50 & 22.81 \\
        20 & 0.1834 & 96.94 & 22.88 & 0.2817 & 94.31 & 23.82 \\
        30 & 0.2240 & 94.47 & 23.85 & 0.3288 & 91.01 & 25.36 \\
        40 & 0.3358 & 88.32 & 25.53 & 0.4487 & 84.43 & 26.72 \\
        50 & 0.5811 & 78.21 & 27.78 & 0.6576 & 76.95 & 27.94 \\
        \midrule
        30 w/ Bypass &  \textbf{0.3425} & \textbf{88.06} & \textbf{25.65} & \textbf{0.4507} & \textbf{84.73} & \textbf{27.09} \\
        \bottomrule
    \end{tabular}
    \label{tab:ablation_no_bypass_step}
\end{table}

\subsection{{Visualization of Bypass $b_t$ under Different $t$}}
\begin{figure}
    \centering
    \includegraphics[width=0.8\linewidth]{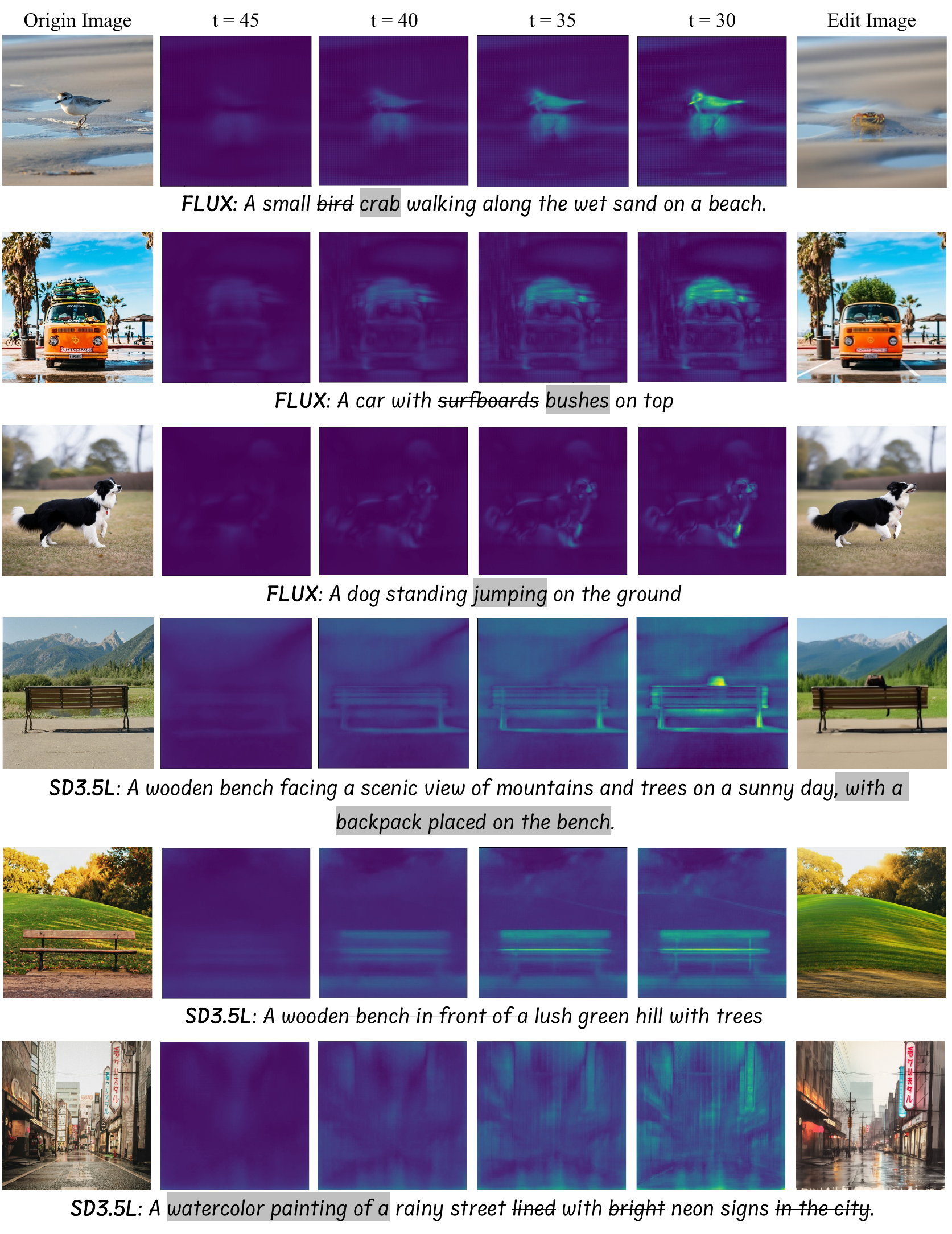}
    \caption{
    {\textbf{Visualization of bypass under different $t_B$.} Zoom in for a better view.}}
    \label{fig:bypass_diff_step_visual}
\end{figure}

{
We present a series of visualizations of bypass $b_t$ under different $t$ during editing.
We would like to clarify that FlowBypass performs only one bypass computation and transition during an actual editing process. This visualization is performed solely to illustrate how the bypass behaves when applied at different bypass timesteps.
As illustrated in Fig.~\ref{fig:bypass_diff_step_visual}, larger values of $t_B$ (\textit{i.e.}, earlier denoising stages) tend to influence global layout and structure, while smaller values of $t_B$ (\textit{i.e.}, later denoising stages) exhibit the modification of local details and texture refinement. This pattern aligns well with observations reported in prior works on DDIM and RF-based sampling, and provides an intuitive view of how the bypass mechanism modulates semantic corrections across the trajectories.
}

\section{{Limitation}}
{Despite its effectiveness, FlowBypass still presents several limitations that highlight opportunities for future improvement.
First, FlowBypass do not show its superiority on editing speed, as computing the bypass term $b_t$ imposes additional overhead. Although FlowBypass is not the fastest method in absolute runtime, it achieves a practical processing cost for high-resolution image editing while offering superior performance and a well-balanced between fidelity and alignment.
Second, FlowBypass shows limited reliability on negation-based prompts (e.g., ``without"). Such edits may fail because the backbone generative backbones inherently struggle with negative conditioning. For instance, prompts such as ``a cat without a hat" often still produce a cat wearing a hat. This issue is shared by many editing methods built upon these backbones, as the models themselves provide weak and unreliable outputs for negation.
Reformulating negation into affirmative phrasing (e.g., ``a cat with a hat" → ``a cat") yields more stable and reasonable results.}


\end{document}